\documentclass[11pt,a4paper]{article}
\usepackage{times,latexsym}
\usepackage{url}
\usepackage[T1]{fontenc}
\usepackage{latexsym}
\usepackage{lscape}
\usepackage[acceptedwithA]{tacl2018v2}
\usepackage{float}
\usepackage{multicol}
\usepackage{multirow}
\usepackage{amssymb}
\usepackage{pifont}

\usepackage{graphicx}
\usepackage{subfigure}
\usepackage{color,colortbl}
\usepackage{xcolor}
\usepackage{microtype}
\definecolor{LightGreen}{HTML}{CCFFCC}
\definecolor{LightYellow}{HTML}{FFF5CC}
\definecolor{LightRed}{HTML}{FFB5B2}
\definecolor{LightBlue}{HTML}{CAEEF7}
\definecolor{Gray}{gray}{0.85}
\newcolumntype{a}{>{\columncolor{Gray}}c}
\newcolumntype{b}{>{\columncolor{white}}c} 

\usepackage{xspace,mfirstuc,tabulary}

\newif\iftaclinstructions
\taclinstructionsfalse 
\iftaclinstructions
\renewcommand{\confidential}{}
\renewcommand{\anonsubtext}{(No author info supplied here, for consistency with
TACL-submission anonymization requirements)}
\newcommand{\instr}
\fi

\iftaclpubformat 

\else

\fi

\title{How To Evaluate Your Dialogue System: Probe Tasks as an Alternative for Token-level Evaluation Metrics}

\author{Prasanna Parthasarathi \textsuperscript{1,2},
  Joelle Pineau \textsuperscript{1,2,3,5},
  Sarath Chandar \textsuperscript{2,4,5} \\
  \textsuperscript{1} School of Computer Science, McGill University \\
  \textsuperscript{2} Quebec Artificial Intelligence Institute (Mila), Canada \\
  \textsuperscript{3} Facebook AI Research (FAIR), Montr\'eal, Canada\\
  \textsuperscript{4} \'Ecole Polytechnique de Montr\'eal \\
  \textsuperscript{5} Canada CIFAR AI Chair
  \\
}

\usepackage{microtype}

\date{}

\begin{document}
\maketitle

\begin{abstract}
Though generative dialogue modeling is widely seen as a language modeling task, the task demands an agent to have a complex natural language understanding of its input text to carry a meaningful interaction with an user. The automatic metrics used evaluate the quality of the generated text as a proxy to the holistic interaction of the agent. Such metrics were earlier shown to not correlate with the human judgement. In this work, we observe that human evaluation of dialogue agents can be inconclusive due to the lack of sufficient information for appropriate evaluation. The automatic metrics are deterministic yet shallow and human evaluation can be relevant yet inconclusive. To bridge this gap in evaluation, we propose designing a set of probing tasks to evaluate dialogue models. The hand-crafted tasks are aimed at quantitatively evaluating a generative dialogue model’s understanding beyond the token-level evaluation on the generated text. The probing tasks are deterministic like automatic metrics and requires human judgement in their designing; benefiting from the best of both worlds. With experiments on probe tasks we observe that, unlike RNN based architectures, transformer model may not be learning to comprehend the input text despite its generated text having higher overlap with the target text.  

\end{abstract}

\section{Introduction}

\protect\citet{kahneman2014thinking} explains decision making with a two system model; an action is decided by either of the two systems -- \emph{System 1} and \emph{System 2}. System 1's decisions are fast and often impulsive -- like pulling one's hand when a vessel is hot, attempting to dodge an object thrown at, etc. In these examples, the system had sufficient information to act quickly and it did so. But, he argues about other scenarios that require higher cognition, where the decision making requires meticulous analysis of facts through a chain of reasoning mechanisms. Such decisions are carried out by System 2. Example scenarios include engaging in debate with a person on say climate change, inflation, politics etc., or a dialogue with a hotel representative to go over a list of holiday options and deciding on the best choice based on price, timing, location and negotiating for deals. Such scenarios require better understanding before action through asking a series of questions. The two systems work differently and require information at different granularity. One of the primary reasons behind it is that, we, humans respond swiftly when the effect of actions is observed immediately and require \textit{additional information} when the decision is not imminent. 

The decision problems in artificial intelligence have variety and can be perceived as a behavior of either of the two systems. From the description and examples cited in \protect\cite{kahneman2014thinking}, Natural Language Processing tasks like
text summarization \protect\cite{luhn1958automatic,kupiec1995trainable,gambhir2017recent},
question answering \protect\cite{rajpurkar2016squad,reddy2019coqa},
sentiment analysis \protect\cite{dave2003mining},
dialogue modeling \protect\cite{vinyals2015neural,serban2015hierarchical,bordes2016learning,li2017adversarial,parthasarathi2018extending,neelakantan2019neural} require careful understanding of the input, as in System 2, to make a decision and act. Other challenging tasks like caption generation \protect\cite{vinyals2015show}, PoS Tagging \protect\cite{klein2003accurate}, machine translation \protect\cite{wang1997decoding,kalchbrenner-blunsom-2013-recurrent}, language modeling \protect\cite{bengio2003neural,rnnlm} require mostly the System 1 dynamics. The task of dialogue modeling requires learning through interaction, often, from humans. The model is expected to understand the input text for it to interact, and the interaction can be meaningful only when the language understanding is better. Approaches for solving dialogue task include information retrieval based approached like selecting a response from a set of canned responses \protect\cite{lowe2015incorporating} or keeping track of very specific information which are \emph{a priori} marked as informative slot-value pairs \protect\cite{guo2018dialog,asri2017frames}; using the current information state to select response. 

Generative dialogue modeling \protect\cite{vinyals2015neural,lowe2015incorporating,serban2015hierarchical, li2016deep,li2017adversarial,parthasarathi2018extending} -- a sub-research in dialogue modeling aims to generate a response as a sequence of tokens with one token at a time conditioned on an input text. The task formulation encapsulates the mechanism for decision making with a free style text generation \protect\cite{vinyals2015neural}. Neural machine translation (NMT) \protect\cite{nmtbahadanau} also followed a similar task formulation and was seeing a dramatic increase in its popularity when generative dialogue models research was getting traction. The successful concepts in NMT got borrowed to other research in NLP like dialogue, text summarization, PoS Tagging which are also sequence mapping problems.

Despite mapping a sequence of text to another text being common between generative dialogue models and NMT, a dialogue generation is required to stay coherent, understand the information in the text, anticipate a response and others in addition to generating text. These underlying mechanisms do not get evaluated with uninformative automatic evaluation metrics like BLEU \protect\cite{papineni2002bleu}, embedding based metrics \protect\cite{wieting2015towards,rus2012optimal,landauer1997solution}, ROUGE \protect\cite{lin2004rouge}, Perplexity Score, METEOR \protect\cite{lavie2007meteor}, F-1 and by training on uninformative dialogue datasets like Ubuntu corpus \protect\cite{lowe2015ubuntu}, Reddit, Twitter \protect\cite{ritter2011data} which do not foster a decision making component with information in addition to sequence generation. To mitigate the requirement of additional information for dialogue generation external knowledge sources like wikipedia \protect\cite{scheepers2017compositionality} and NELL \protect\cite{carlson2010toward} was tried with reasonable success in dialogue generation \protect\cite{parthasarathi2018extending,dinan2018wizard}. 

But, the issues in evaluation -- automatic evaluation metrics uncorrelated with human judgement -- showcased by \protect\citet{liu2016not} is still an open problem. Attempts to mimic human scores for better evaluation metric \protect\cite{lowe2017towards} and other metrics that aim to correlate to the human judgement \protect\cite{sinha2020learning,tao2018ruber} still have a long way to go. Dependency on human evaluation is not any better as it is subjective, requires more time, cost, effort and has issues of disagreement among scorers in evaluation \protect\cite{li2019acute}. But, many dialogue research promote the costly yet inconclusive human evaluation as a way for thorough evaluation of dialogue models. The evaluation of surface level token generation -- automatic or with a human -- does not evaluate the underlying understanding of a dialogue model, which is required of a model to have meaningful conversation. The language understanding component of an agent more often than not goes unnoticed with only token-level evaluation metrics.

Simpler probing tasks \protect\cite{belinkov2019analysis,jawahar2019does,anand2019unsupervised} test the performance of classifiers that train on the representation learnt by a model. A probe task is a backward reasoning task, where a model reasons out its understanding of the input through simple questions as classification task. It could be seen that probe tasks can evaluate whether the model does careful reasoning on the facts presented as like in \emph{System 2}. For example, in representation learning for reinforcement learning, learning to predict the position of an enemy from the encoding of input state shows whether the representation is discriminative of the game features. The probe tasks allow a way to quantify the understanding of a model and articulate a meaningful discussion around the success or failure of a model, prevent over-fitting to spurious patterns, identify unwanted biases in a model among others. Drawing inspirations and analysing the existing literature to the best of our knowledge we propose a set of probing tasks\footnote{The code repository can be found at \href{https://github.com/ppartha03/Dialogue-Probe-Tasks-Public}{https://github.com/ppartha03/Dialogue-Probe-Tasks-Public}} for evaluating the language understanding of generative dialogue models on chit-chat and goal-oriented dialogues --  (Table \ref{tab:probe-tasks-list}). Our probe tasks also help us to understand the difference in learning behaviour of recurrent neural network (RNN) models and Transformer models which was previously not evident from the token-level evaluation methods.

Our contributions in the paper are:
\begin{itemize}
    \item Showcasing the significantly high variance in human evaluation of dialogues.
    \item Proposing a list of probe tasks -- 2 semantic, 13 information specific and 3 downstream as an alternate evaluation of dialogue systems.
    \item Finding that the representation learnt by recurrent neural network based models is better at solving the probe tasks than the one by transformer model.
\end{itemize}
\section{Related Work}
\subsection{Language Generation}
Setting up probe-tasks to understand the underlying workings of the neural models is not unique. There has been quite a lot of work \protect\cite{conneau2018you,belinkov2019analysis,elazar2020bert} to understand the embedded information in vector embedding of sentences. The objective of designing probe tasks is to evaluate the inductive bias of a model to learn a task-specific hidden representation that can be used to solve a series of simple tasks. As it is easier to control the biases in probing tasks than in the downstream tasks, research in language generation has analysed models on probing tasks like using encoder representation to identify words in input (\textbf{WordCont}) to measuring encoder sensitivity to shifts in bigrams \protect\cite{conneau2018you,belinkov2019analysis}.
Dialogue generation task requires understanding and reasoning on the information from a user before generating text and language generation is only a part of the task. The probing tasks on language generation are useful in probing the generation aspect of dialogue models but are not complete to measure the dialogue modeling capabilities of models, which involves the agent having a better understanding of the input context.
\subsection{Reinforcement Learning}
Reinforcement learning (RL) is another area of research where large models make it hard to deduce whether a model is biased in the right ways to act in an environment. Dialogue generation shares similarity with the sequential action prediction in RL by their large combined space of possible states and actions. To avoid learning from spurious correlation in the data, it is important to verify if the agents are getting biased by relevant features in input. \protect\citet{anand2019unsupervised} learn state representation for an RL agent in an unsupervised setting and introduce a set of probing tasks to evaluate the representation learnt by agents. This includes using an annotated dataset with markers for position of the agent, current score, items in inventory, target's location among others. The authors train a shallow linear classifier and measure its performance; which serves as a metric for the representational soundness of the learning algorithm. Due to the similarity between RL and dialogue, we draw inspirations from \protect\citet{anand2019unsupervised}'s probing tasks on game playing agent.
\subsection{Probing tasks in Vision}
Applications of computer vision like caption generation for images \protect\cite{vinyals2015show} or videos \protect\cite{donahue2015long} use attention based models to parse over the hidden states of a convolutional neural network (ConvNet) \protect\cite{lecun1998gradient}. The attention over the ConvNet features are visualized to observe the words corresponding to different parts of the image. Visualizing the attention has been one of the qualitative probe task for text generation conditioned on images \protect\cite{xu2015show}.
\subsection{Software Unit Testing}
It is also interesting to draw parallels to \emph{Unit Testing} in software engineering \protect\cite{koomen1999test}, where the smallest software components of a system are tested for their design and logical accuracy. The only difference between a deterministic application software and a stochastic decision making ML module is that the behavior of the ML system is data-driven while for a software system it is driven by logic. Despite the difference, the unit testing and probing tasks share a common ground. Analogous to an application software, the decision making modules also have smaller decision components. In most models, these smaller components are latent but still they have to be evaluated as they indirectly contribute to the performance of the modules. These latent components can be explicitly validated to qualify a model's understanding and to keep a check on irrelervant and/or irresponsible biases the agent may have picked up from the dataset. Such evaluations on ML systems will have positive effect on users' trust when they are deployed.
\section{Dialogue Probing Tasks}
Dialogue is a complex decision making problem, which is sequential and requires the agent to have sufficient understanding of the context before generating a response. The quality of the generated text is not a sufficient metric to evaluate the model's understanding of the input. An evaluation metric for dialogue requires a profound understanding of grammar, semantics, as well as domain specific and general knowledge; this makes the problem of coming up with the evaluation metric for dialogue AI-Complete \protect\cite{yampolskiy2013aicomplete}. In a way, if we have a perfect evaluation metric, the metric itself is an AI system which explains the rationale behind the need for human evaluation. But, human evaluation is difficult to scale up, and agreement among annotators is hard to come by, and formulating the right questions takes effort \protect\cite{li2019acute}; so, instead of spending human effort in appropriately evaluating the dialogue, we propose to design probe tasks -- semantic, syntactic, information specific and downstream tasks -- for each dialogue dataset. The tasks proposed and discussed in this paper are shown in Table \ref{tab:probe-tasks-list}. 


\subsection{Semantic and Syntactic Probe Tasks} Similar to probing a language generation model, some probe tasks for dialogue generation models include measuring sensitivity of models to context by shuffling the input \protect\cite{chinnapaper}, testing if the model can predict a mid frequency token in the context (\textbf{WordCont})\protect\cite{belinkov2019analysis}, and testing if the model understands how far is it in the conversation by using its context encoding (\textbf{UtteranceLoc}) \protect\cite{sinha2020learning} . To formally compare different dialogue generation models, we use \textbf{UtteranceLoc} and \textbf{WordCont} probe tasks to evaluate the semantic understanding. 

\subsection{Information Specific Probe Tasks} Apart from language generation, a dialogue agent is expected to understand specific information from the context to help it in generating the appropriate response. The surface level observation of "\textit{nice human like text}" or other syntactic or semantic features do not shed light on the underlying mechanisms. Hence, it is imperative that we systematically probe also the information processing with simple questions on agent's encoding of the input context. To this end, we propose 12 information specific probe tasks that track the understanding of a dialogue model on the input text. The tasks are listed in Table \ref{tab:probe-tasks-list}.
\begin{table}[!h]
    \centering
    \tiny
    \begin{tabular}{|p{1cm}|l|p{4.2cm}|}
     \hline
     \textbf{Task} & \textbf{Task Name} & \textbf{Description} \\
     \hline
      \multirow{2}{1cm}{Semantic}&\cellcolor{LightYellow}\textbf{UtteranceLoc}$^{*}$&\cellcolor{LightYellow} How long has the conversation been happening ?\\
      &\cellcolor{LightYellow}\textbf{WordCont}$^{+}$ & \cellcolor{LightYellow}Which mid-frequency word is encoded in the context ? \\
      \hline
       \multirow{13}{1cm}{Information Specific}&\cellcolor{LightGreen}\textbf{IsMultiTopic} & \cellcolor{LightGreen}Does the conversation have more than one topic ? \\
       &\cellcolor{LightGreen}\textbf{NumAllTopics} &\cellcolor{LightGreen} How many topics does this conversation have ? \\
       &\cellcolor{LightGreen}\textbf{RepeatInfo} & \cellcolor{LightGreen}Which information provided by the user is repeated ?\\
       &\cellcolor{LightGreen}\textbf{NumRepeatInfo} & \cellcolor{LightGreen}What many number of recent information are repeats ? \\
       &\cellcolor{LightYellow}\textbf{AllTopics}  & \cellcolor{LightYellow}What are all the topics discussed so far ? \\
       &\cellcolor{LightYellow}\textbf{RecentSlots} &\cellcolor{LightYellow} What is the \textit{recent} information given by the user ? \\
       &\cellcolor{LightYellow}\textbf{NumRecentInfo} &\cellcolor{LightYellow} How \emph{many} information did the user provide \textit{recently} ?\\
       &\cellcolor{LightRed}\textbf{RecentValues} &\cellcolor{LightRed} What are the details of the \textit{recent} information ? \\
       &\cellcolor{LightRed}\textbf{AllSlots} &\cellcolor{LightRed} What \textit{all} information are given by the user so far? \\
       &\cellcolor{LightRed}\textbf{AllValues} & \cellcolor{LightRed}What are the details of in \textit{all} the information ? \\
       &\cellcolor{LightRed}\textbf{RecentTopic} &\cellcolor{LightRed} What is the current topic of the dialogue ? \\
       &\cellcolor{LightRed}\textbf{NumAllInfo} &\cellcolor{LightRed}How \emph{many} information did the user provide so far ?\\
       &\cellcolor{LightRed}\textbf{PersonalInfo}$^{+}$ &\cellcolor{LightRed}What characteristics of your persona can you infer from the conversation history ?\\
     \hline
     \multirow{3}{1cm}{Downstream task}&\cellcolor{LightRed}\textbf{ActionSelect}&\cellcolor{LightRed} Which downstream task (database query) follows the current conversation ? \\
       &\cellcolor{LightRed}\textbf{EntitySlots} &\cellcolor{LightRed}What information should is required to construct the query ? \\
       &\cellcolor{LightRed}\textbf{EntityValues} &\cellcolor{LightRed}What values should be passed to the query ?\\
       \hline
       
    \end{tabular}
    \caption{List of probe tasks on the two different datasets. The color of the row identifies its difficulty -- green, yellow and red for easy, medium and hard respectively. This categorization is further discussed in Section \ref{sec:discussion}.
    $^{+}$ - Task on PersonaChat, $^{*}$ - Task on both the datasets, 'none' - Task on MultiWoZ.}
    \label{tab:probe-tasks-list}
\end{table}
\subsection{Downstream Probing Tasks}
A Dialogue agent is not only expected to predict the next utterance, and track the information but also perform an appropriate downstream task. An example could be of an agent carrying out a follow-up action \texttt{Hotel-recommend} with \texttt{guesthouse-price} and \texttt{cheap} as slot and value to a user context -- "\emph{I am looking for a low-price hotel to stay.}"  The dialogue state tracking measures the performance of a model on such tasks \protect\cite{henderson2014second} but such often goes unnoticed in the evaluation of generative models. Works like that of \protect\citet{neelakantan2019neural} use entity, values and action information to train on the dialogue generation task but the performance of a generative dialogue model without explicitly training on the downstream tasks are not compared. Towards that, we propose \textbf{ActionSelect}, \textbf{EntitySlots}, \textbf{EntityValues} probe tasks. The details of the task are shown in Table \ref{tab:probe-tasks-list}.

\section{Experiments}
\subsection{Datasets}
We experiment the proposed set of probing tasks on MultiWoZ 2.0 \protect\cite{budzianowski2018multiwoz} -- with goal-oriented dialogues and PersonaChat \protect\cite{zhang2018personalizing} -- with chit-chat dialogues. The features of the datasets are shown in Table \ref{tab:dataset}. The data sets represent the two major styles in dialogue and we use probe tasks accordingly.

For experiments in this paper, we used BLEU score \protect\cite{papineni2002bleu} on the validation set as the metric for model selection, which is commonly used in dialogue research. The results of the experiments did not differ with using other model selection metrics like ROUGE-F1, METEOR or Vector-Based (Average BERT embedding) which are tabled and explained in Appendix \ref{sec:addn-experiments}.
\begin{table}[h]
    \centering
    \small
    \begin{tabular}{c|c|c|c}
    \hline
    {\bf Dataset} & {\bf Train } & {\bf Validation} & {\bf Vocabulary} \\
    \hline
         {\it PersonaChat} & $\sim$ 10900 & 1500 & 16k \\
         {\it MultiWoZ} & $\sim$ 8400 & 1000 & 13k \\
    \hline
    \end{tabular}
    \caption{Distribution of the dialogues in PersonaChat and MultiWoZ.}
    \label{tab:dataset}
\end{table}
\subsection{Models}
We train 5 commonly used generative dialogue models for 25 epochs on the two datasets.  

\paragraph{\textsc{LSTM Encoder-Decoder}} The architecture \protect\cite{vinyals2015neural} has an LSTM cell to encode the input context only in the forward direction. For a sequence of words in the input context $(w_1^i, w_2^i, \ldots, w_{T'}^i)$ LSTM encoder generates $\{h_t\}_1^T$. The decoder LSTM's hidden state is initialized with $h_t^T$. The decoder outputs one token every step. We used two layer LSTM cell; the first layer applies recurrent operation on the input to the model while the layer above recurs on the outputs of the layer below. The encoder final hidden state (from the 2nd layer) is passed as an input to the decoder. We train the model with cross entropy loss as shown in Equation \ref{eqn:cross-entropy}.
\begin{equation}
    \sum_{t=1}^{T} -y_t \log (p\left(\hat{y}_t)\right) - (1-y_t) \log (1- p\left(\hat{y}_t)\right) 
    \label{eqn:cross-entropy}
\end{equation}
where $y_t$ is the $t^{th}$ ground truth token distribution in the output sequence, $\hat{y}_t$ is model generated token and $p$ is the model learned distribution over the tokens.
We train the model with Adam \protect\cite{kingma2014adam} optimizer with teacher forcing \protect\cite{teacherforcing}.
\paragraph{\textsc{LSTM Encoder-Attention Decoder}} The architecture is similar to the LSTM Encoder-Decoder with an exception of an attention module to the decoder. The attention module \protect\cite{nmtbahadanau} linearly combines the encoder hidden states ${h_t}_1^{T}$ as an input to the decoder LSTM at every step of decoding, unlike only having the last encoder hidden state.
\begin{table}[h!]
    \centering
    \small
    \begin{tabular}{p{5cm}|c}
    \hline
       {\bf Model}  & {\bf Parameters}  \\
    \hline
        \rowcolor{Gray}{\it LSTM Encoder-Decoder} &	11M \\
        {\it LSTM Encoder-Decoder + Attention} &	11M \\
        \rowcolor{Gray}{\it HRED} &	12M \\
        {\it Bi-LSTM Encoder-Decoder}	& 12M\\
        \rowcolor{Gray}{\it Transformer}	& 41M \\
    \hline
    \end{tabular}
    \caption{Size of parameters of the models used in all the experiments on the two datasets. \emph{M} for Million.}
    \label{tab:size-of-models}
\end{table}
\paragraph{\textsc{Hierarchical Recurrent Encoder Decoder}} The model has encoding done by two encoder modules acting at different levels \protect\cite{sordoni2015hierarchical}; \textit{sentence encoder} to encode the sentences that feeds in as input to the \textit{context encoder}. Both the encoders are LSTMs. The decoder is an attention decoder.
\paragraph{\textsc{Bi-LSTM Encoder-Attention Decoder}} The encoder is a concatenation of two LSTMs that can read the input from forward and backward direction \protect\cite{bidirectional}. The hidden state is computed as the summation of the hidden states of the two encoders. The decoding is done with an attention decoder.
\paragraph{\textsc{Transformer Architecture}} This state-of-the-art architecture \protect\cite{vaswani2017attention,rush2018annotated} is a transductive model that has multiple layers of attention to predict the output. We used the architecture in an encoder-decoder style by splitting half the layers for encoding and the remainder for decoding. We perform the probe tasks on the encoder hidden state.

The size of the models used in the experiments are detailed in Table \ref{tab:size-of-models}. For the probing tasks, we select the untrained model, model with the best BLEU score on validation, and model from the last training epoch. 
We use packages pytorch \protect\cite{paszke2017pytorch} and scikit-learn \protect\cite{scikit-learn} for our experiments.

\subsection{Motivation for Dialogue Probe Tasks}
\label{sec:motivation}
Although criticism on automatic metrics for dialogue evaluation \protect\cite{liu2016not,chinnapaper} is widely accepted, the human evaluation, though straightforward, does not validate the holistic understanding of the models being compared against. Successful evaluation requires careful crafting of the appropriate question and is contingent on the understanding of the same by the human participants. Here, the evaluation expects a human participant to understand not only a model's ability to generate meaningful text but also verify if it understood the conversation thus far; which requires more details for the human participants to have an agreement. We hypothesize that the generated text presented to the participants are greatly dependent on the choice of seed values and a slight variation could result in a model generating a very different response. We verify that the human participants cannot identify the difference between two models by posing an alternate hypothesis where we expect the participants to fail in rating alike the two responses selected from two different runs of the same model with different seed values.

\begin{figure}[h!]
    \centering
    \includegraphics[width=0.8\columnwidth]{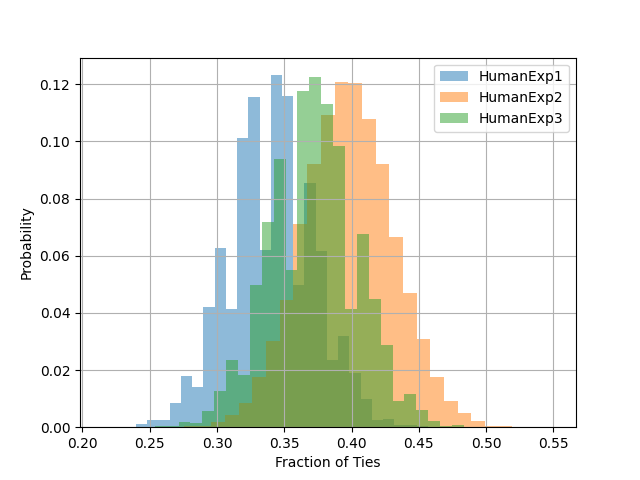}
    \caption{The mean of the distribution of tie in three different experiments was centered around 35\%, showing that the subjective scores on responses by humans are not sufficient to evaluate a model. }
     \label{fig:human_bias_goal}
\end{figure}

To verify our hypothesis, we train the models on the two datasets and performed human evaluation experiment with 500 volunteers through amazon's Mechannical Turk \protect\cite{buhrmester2016amazon,miller2017parlai}. For an even more challenging scenario to validate our hypothesis, we chose the goal-oriented dataset as the context-response variance is relatively lower than in PersonaChat. 
\begin{table}[h!]
    \centering
    \small
    \begin{tabular}{c|c|c}
    \hline
         {\bf Model}& {\bf PersonaChat} & {\bf MultiWoZ}  \\
    \hline
        \rowcolor{Gray}{\it BiLSTM + Attn} & 4.4 $\pm$ 0.06 & 15.5 $\pm$ 0.05 \\
        {\it Seq2Seq} & 4.5 $\pm$ 0.06 & 15.8 $\pm$ 0.17 \\
        \rowcolor{Gray}{\it Seq2Seq + Attn} & 4.4 $\pm$ 0.15 & 15.7 $\pm$ 0.11\\
        {\it HRED} & 3.9 $\pm$ 0.01 & 12.2 $\pm$ 4.00 \\
        \rowcolor{Gray}{\it Transformer} & 7.9 $\pm$ 0.17 & 29.4 $\pm$ 0.61\\
    \hline
    \end{tabular}
    \caption{BLEU scores of the models from runs with different seeds on PersonaChat and MultiWoZ dataset. (Higher the better. We measure BLEU-2 (case insensitive).}
    \label{tab:bleu-scores}
\end{table}
For the study, we sample 2000 context-response pairs from Bi-LSTM Attention model from two different seeds. We chose this model as this had the lowest variance in BLEU scores (Table \ref{tab:bleu-scores}). We ask the participants to select the response that they think is more \emph{relevant} to the given context, similar to \protect\citet{li2015diversity}. The annotators can select either of the responses or a Tie. We show the participants the responses generated by the model with same model parameters, but different seeds. For every context-response pair, we collected 3 feedback from different participants (Distribution corresponding to the 3 different human responses are shown with legend HumanExp1, HumanExp2 and HumanExp3 in Figure \ref{fig:human_bias_goal}). 
Usually human evaluation is done on 100-500 responses. To understand the variance in this set up and the lack of information at the token generation level, we sample 50000 sets of 200 human responses from the collected 2000 responses and compute the fraction of times there was a tie. We observed that distribution over the fraction of times the human participants selected a \textit{Tie} was centered around 35\% (Figure \ref{fig:human_bias_goal}) with all of the probability mass within 50\%. This shows that (a) text generated by the same model can have variance with the seeds, and the variance is significant (b) attributing the choice of seed value to the performance of a model creates confusion in the evaluation. The results show that the scores based only on the text generated by a model cannot be extrapolated to be the performance of the model architecture in the dialogue task. Optimizing on the seed value does not guarantee reproducible results, which is necessary to progress the field further. Further, the dependence of the model generated text on seed value raises a valid concern; whether a model parameter chosen by the seed value can mimic the surface level token generation of a model that actually understands the context. The lack of clarity leads to inconclusiveness of studies with human evaluation only on the generated text.

Although variance due to the seeds can be reduced with averaging results over multiple seed values, human evaluations are expensive and requires the same effort for setting them up every time. Whereas probe-tasks allow cheaper extension for evaluating over multiple seed values that can effectively reduce the variance on an appropriate set of probe tasks.

\subsection{Probing Tasks}
We train the models on the two different datasets without any auxiliary tasks. To understand the evolution on the probe task from beginning of the training till end, we compare with 3 different parameter configurations of every model -- \emph{Untrained}, \emph{Last epoch}, and \emph{BestBLEU}. We save the model parameters while training on the end-to-end dialogue generation task and evaluate on the probing tasks as a post analysis. We use Logistic Regression classifier\footnote{We also trained a nonlinear model --multi-layer perceptron for probe tasks (Appendix \ref{sec:addn-experiments}). The results had a similar trend holding good the remainder of the discussion in the paper.} implementation from scikit-learn \protect\cite{scikit-learn} with default parameters except the max\_iter set to 250 for all the probing tasks, invariably. We train the classifier using the encoder representations on the probe tasks with the training set and evaluate with the validation set. The evaluation metric is \emph{F1}-score with micro averaging in multi-class prediction tasks. The data preparation for the probing tasks are discussed in detail in Appendix \ref{sec:data-preparation}.
\paragraph{\textsc{Probe Tasks on PersonaChat}} The models are evaluated on the three probing tasks relevant to chit-chat dialogue generation (Table \ref{tab:persona-probe-tasks}) -- two semantic and one information specific. {\it UtteranceLoc} and {\it WordCont} measures if the encoded context suggests semantic awareness of the model while \textit{PersonalInfo} measures the amount of knowledge the model has about its persona from encoding of conversation history. In other words, it evaluates the extent to which persona can be identified from the context encoding with a linear classifier. A better performance in these tasks indicate that the model has an understanding that conversations involve assuming different persona, utterances follow a temporal sequence and hence the encoding has to be different. Would a human have such an understanding ? Yes. 

The PersonalInfo task here is not very specific to identifying personal information but acts as an indicator to the information embedded in dialogues that goes unnoticed in the encoding. It was surprising to see that none of the models scored a reasonable \textit{F1}. Transformer model scored higher on BLEU score (Table \ref{tab:bleu-scores}) but performance of transformer on PersonalInfo task was decreasing with increased training (Table \ref{tab:persona-probe-tasks}).

\begin{table}[h!]
\tiny
\centering
      \begin{tabular}{c|a|b|a}
         \hline
         \multicolumn{4}{c}{PersonaChat Dataset}\\
         \hline
         {\bf Model} & {\bf UtteranceLoc} & {\bf WordCont} & {\bf PersonalInfo} \\
         \hline
         \rowcolor{LightBlue}\multicolumn{4}{c}{Bi-LSTM Seq2Seq + Attention} \\
         \hline
         Untrained &36.97 $\pm$ 0.12 &43.53 $\pm$ 0.02 &0.00 $\pm$ 0.00    \\
         LastEpoch & 56.53 $\pm$ 0.02 &39.93 $\pm$ 0.04 &0.03 $\pm$ 0.00   \\
         BestBLEU & 57.19 $\pm$ 0.05 &39.72 $\pm$ 0.08 &0.02 $\pm$ 0.00   \\
        \hline
         \rowcolor{LightBlue}\multicolumn{4}{c}{HRED - LSTM} \\
         \hline
         Untrained & 1.17 $\pm$ 0.02 &51.73 $\pm$ 0.02 &0.00 $\pm$ 0.00 \\
         LastEpoch &12.81 $\pm$ 4.93 &49.42 $\pm$ 0.29 &0.00 $\pm$ 0.00   \\
         BestBLEU & 10.76 $\pm$ 3.48 &51.00 $\pm$ 0.07 &0.00 $\pm$ 0.00  \\
         \hline
         \rowcolor{LightBlue}\multicolumn{4}{c}{LSTM Seq2Seq + Attention} \\
         \hline
         Untrained & 39.92 $\pm$ 0.04 &47.19 $\pm$ 0.05 &0.00 $\pm$ 0.00 \\
         LastEpoch & 51.98 $\pm$ 0.02 &39.97 $\pm$ 0.02 &0.00 $\pm$ 0.00  \\
         BestBLEU & 54.06 $\pm$ 0.06 &43.77 $\pm$ 0.24 &0.00 $\pm$ 0.00  \\
         \hline
         \rowcolor{LightBlue}\multicolumn{4}{c}{LSTM Seq2Seq} \\
         \hline
         Untrained & 40.19 $\pm$ 0.02 &46.91 $\pm$ 0.03 &0.00 $\pm$ 0.00   \\
         LastEpoch & 50.85 $\pm$ 0.14 &39.98 $\pm$ 0.00 &0.04 $\pm$ 0.00    \\
         BestBLEU & 52.23 $\pm$ 0.08 &40.16 $\pm$ 0.04 &0.01 $\pm$ 0.00  \\
         \hline
         \rowcolor{LightBlue}\multicolumn{4}{c}{Transformer Architecture} \\
         \hline
         Untrained & 52.96 $\pm$ 0.01 &35.91 $\pm$ 0.01 &2.35 $\pm$ 0.00  \\
         LastEpoch & 42.65 $\pm$ 0.11 &46.86 $\pm$ 0.09 &0.00 $\pm$ 0.00   \\
         BestBLEU & 40.73 $\pm$ 0.06 &46.16 $\pm$ 0.03 &0.03 $\pm$ 0.00   \\
         \hline
    \end{tabular}
    \caption{Performance of different models on the probe tasks on PersonaChat dataset. The performance is measured as \emph{F-1} score (Higher the better).}
    \label{tab:persona-probe-tasks}
\end{table}

The semantic tasks UtteranceLoc and WordCont evaluate whether the model understands how far in the conversation is it in and if it can identify mid-frequency words in the target response. Bi-LSTM model performed the best in UtteranceLoc while the Transformer model was not in the top 3. Transformer performed the best in WordCont. 

We hypothesize that the Transformer model can learn extensive specific information in the input because of the size of its attention but finds it difficult to learn general information like in UtteranceLoc. Also, we observed that the inductive biases of the SEQ2SEQ models enable random projections that are informative even without training. This correlates with independent observations on the results in \protect\cite{worldmodelblog} which suggests random projections on temporal information can hold information. Similarly, Transformer architecture's random representation is also informative; but visualization of the encoder hidden states in low dimensional space does not show prominent cluster formations unlike in SEQ2SEQ -- Figure \ref{fig:personachat-context} in Appendix \ref{sec:addn-experiments}. The SEQ2SEQ models have a smaller manifold due to recurrent multiplication that regularizes its representation to observe structures, whereas Transformer network's attention operations project the context on to a larger manifold that prevents loss in encoding \footnote{\protect\citet{ramsauer2020hopfield} showed recently that the transformer model is a large look-up table. Our empirical results support the authors' view.}. This explains the SEQ2SEQ models performing well on UtteranceLoc while Transformer model performing well on WordCont. The difference between the two classes of models is much more evident on the probing tasks in MultiWoZ dataset.

\paragraph{\textsc{Probe Tasks on MultiWoZ}} Unlike chit-chat dialogues, goal-oriented datasets naturally provide probe-tasks that can validate the understanding of a model on the task. The probe tasks as shown in Table \ref{tab:probe-tasks-list} enable the hidden representation of an end-to-end goal-oriented dialogue agent to be interpretable. We tested the models on 16 different probe tasks -- 1 Semantic, 12 Information specific and 3 downstream tasks. These probe tasks are designed to reveal the models' ability to understand information in conversation history only through indirect signal -- language generation. 

In majority of information specific tasks and in the downstream tasks (Tables \ref{tab:multiwoz-probe-tasks-1},\ref{tab:multiwoz-probe-tasks}), we observed that SEQ2SEQ models performed significantly better than the Transformer model. Interestingly, we observed a pattern in Transformer in the two datasets, that the model's performance on the probe tasks decreased from the beginning of training till the end on all of the tasks, while for the rest of the models there was learning involved. 

  \begin{table*}[h!]
        \tiny
        \centering
        \begin{tabular}{c|a|b|a|b|a|b|a|b}
         \hline
         \multicolumn{9}{c}{MultiWoZ Dataset}\\
         \hline
         {\bf Model} & {\bf UtteranceLoc} & {\bf RecentTopic} & {\bf RecentSlots} & {\bf RecentValues} & {\bf RepeatInfo} & {\bf NumRepeatInfo}& {\bf NumRecentInfo} & {\bf AllSlots} \\
         \hline
         \rowcolor{LightBlue}\multicolumn{9}{c}{LSTM Seq2Seq + Attention} \\
         \hline
         Untrained &32.67 $\pm$ 0.71 &18.97 $\pm$ 0.01 &30.01 $\pm$ 0.06 &25.31 $\pm$ 0.04 &71.31 $\pm$ 0.00 &75.22 $\pm$ 0.00 &35.85 $\pm$ 0.03 & 9.91 $\pm$ 0.03 \\
         LastEpoch &57.00 $\pm$ 0.05 &88.88 $\pm$ 0.00 &65.47 $\pm$ 0.01 &40.60 $\pm$ 0.02 &70.15 $\pm$ 0.00 &74.45 $\pm$ 0.01 &60.86 $\pm$ 0.02 &52.56 $\pm$ 0.02\\
         BestBLEU & 57.55 $\pm$ 0.05 &89.91 $\pm$ 0.07 &67.39 $\pm$ 0.02 &40.49 $\pm$ 0.04 &70.92 $\pm$ 0.00 &74.73 $\pm$ 0.00 &62.48 $\pm$ 0.02 &53.08 $\pm$ 0.11 \\
        \hline
         \rowcolor{LightBlue}\multicolumn{9}{c}{HRED - LSTM} \\
         \hline
         Untrained & 10.82 $\pm$ 0.15 &0.00 $\pm$ 0.00 &26.64 $\pm$ 0.03 & 22.83 $\pm$ 0.02 &72.15 $\pm$ 0.01 &76.01 $\pm$ 0.00 &35.14 $\pm$ 0.03 &0.00 $\pm$ 0.00 \\
         LastEpoch & 37.22 $\pm$ 10.39 &54.22 $\pm$ 22.71 &36.88 $\pm$ 10.51 &20.92 $\pm$ 3.29 &71.39 $\pm$ 0.00 &74.81 $\pm$ 0.01 &39.03 $\pm$ 11.42 &31.77 $\pm$ 8.04 \\
         BestBLEU & 37.15 $\pm$ 10.35 &50.98 $\pm$ 20.94 &34.84 $\pm$ 9.69 &20.63 $\pm$ 3.21 &71.68 $\pm$ 0.00 &75.06 $\pm$ 0.00 &38.59 $\pm$ 11.18 &30.23 $\pm$ 7.84 \\
         \hline
         \rowcolor{LightBlue}\multicolumn{9}{c}{LSTM Seq2Seq } \\
         \hline
         Untrained &32.40 $\pm$ 0.87 &18.57 $\pm$ 0.03 &30.45 $\pm$ 0.01 &25.01 $\pm$ 0.01 &71.64 $\pm$ 0.00 &75.48 $\pm$ 0.00 &35.82 $\pm$ 0.01 &10.66 $\pm$ 0.01  \\
         LastEpoch &56.31 $\pm$ 0.04 &88.45 $\pm$ 0.01 &65.93 $\pm$ 0.00 &39.46 $\pm$ 0.02 &70.05 $\pm$ 0.02 &73.94 $\pm$ 0.02 &60.38 $\pm$ 0.01 &52.34 $\pm$ 0.03\\
         BestBLEU& 57.37 $\pm$ 0.06 &89.45 $\pm$ 0.03 &68.08 $\pm$ 0.01 &39.78 $\pm$ 0.07 &71.28 $\pm$ 0.01 &75.36 $\pm$ 0.01 &62.33 $\pm$ 0.05 &53.40 $\pm$ 0.05 \\
         \hline
         \rowcolor{LightBlue}\multicolumn{9}{c}{Bi-LSTM Seq2Seq + Attention} \\
         \hline
         Untrained & 38.98 $\pm$ 0.04 &23.20 $\pm$ 0.02 &28.56 $\pm$ 0.07 &23.82 $\pm$ 0.03 &71.61 $\pm$ 0.01 &75.58 $\pm$ 0.00 &34.45 $\pm$ 0.02 &17.96 $\pm$ 0.00\\
         LastEpoch & 57.12 $\pm$ 0.04 &87.16 $\pm$ 0.02 &63.60 $\pm$ 0.00 &38.28 $\pm$ 0.01 &70.71 $\pm$ 0.00 &74.70 $\pm$ 0.01 &59.02 $\pm$ 0.03 &52.94 $\pm$ 0.01  \\
         BestBLEU & 59.04 $\pm$ 0.10 &89.85 $\pm$ 0.03 &65.03 $\pm$ 0.00 &39.06 $\pm$ 0.00 &71.98 $\pm$ 0.01 &75.63 $\pm$ 0.00 &60.36 $\pm$ 0.05 &54.96 $\pm$ 0.05  \\
         \hline
         \rowcolor{LightBlue}\multicolumn{9}{c}{Transformer Architecture} \\
         \hline
         Untrained &46.74 $\pm$ 0.03 &80.09 $\pm$ 0.03 &46.12 $\pm$ 0.02 &36.11 $\pm$ 0.00 &72.25 $\pm$ 0.01 &75.05 $\pm$ 0.01 &48.00 $\pm$ 0.01 &65.20 $\pm$ 0.01 \\
         LastEpoch & 37.86 $\pm$ 0.05 &49.73 $\pm$ 0.75 &28.50 $\pm$ 0.06 &23.79 $\pm$ 0.02 &71.99 $\pm$ 0.00 &75.72 $\pm$ 0.00 &37.54 $\pm$ 0.01 &37.18 $\pm$ 0.55\\
         BestBLEU &39.46 $\pm$ 0.00 &57.05 $\pm$ 1.50 &30.10 $\pm$ 0.27 &23.72 $\pm$ 0.03 &72.70 $\pm$ 0.00 &75.97 $\pm$ 0.00 &39.11 $\pm$ 0.08 &40.43 $\pm$ 1.21\\
         \hline
    \end{tabular}
    \caption{The performance of different generative dialogue models on probe tasks in MultiWoZ dialogue data set. The performance is measured with \emph{F1} (Higher the better). The results show that SEQ2SEQ models perform significantly better than Transformer model on the probe tasks, despite the models falling behind in BLEU score.}
    \label{tab:multiwoz-probe-tasks-1}
\end{table*}

\begin{table*}[h!]
\tiny
\centering
    \begin{tabular}{c|a|b|a|b|a|b|a|b}
         \hline
         \multicolumn{9}{c}{MultiWoZ Dataset}\\
         \hline
         {\bf Model} & {\bf AllValues} & {\bf NumAllInfo} & {\bf AllTopics}&{\bf NumAllTopics} & {\bf IsMultiTask} & {\bf EntitySlots} & {\bf EntityValues} & {\bf ActionSelect} \\
         \hline
         \rowcolor{LightBlue}\multicolumn{9}{c}{LSTM Seq2Seq + Attention} \\
         \hline
         Untrained & 2.91 $\pm$ 0.00 &0.00 $\pm$ 0.00 &35.98 $\pm$ 0.00 &77.26 $\pm$ 0.01 &85.13 $\pm$ 0.01 &20.12 $\pm$ 0.02 &17.88 $\pm$ 0.01 &14.99 $\pm$ 0.00  \\
         LastEpoch & 13.40 $\pm$ 0.00 &25.77 $\pm$ 0.04 &74.45 $\pm$ 0.02 &78.80 $\pm$ 0.02 &84.43 $\pm$ 0.00 &41.79 $\pm$ 0.01 &30.89 $\pm$ 0.02 &58.50 $\pm$ 0.01\\
         BestBLEU  &12.81 $\pm$ 0.01 &25.73 $\pm$ 0.02 &75.33 $\pm$ 0.02 &79.39 $\pm$ 0.02 &85.30 $\pm$ 0.00 &41.29 $\pm$ 0.03 &31.57 $\pm$ 0.03 &60.14 $\pm$ 0.01\\
        \hline
         \rowcolor{LightBlue}\multicolumn{9}{c}{HRED - LSTM} \\
         \hline
         Untrained & 0.00 $\pm$ 0.00 &0.00 $\pm$ 0.00 &23.00 $\pm$ 0.03 &77.40 $\pm$ 0.01 &85.07 $\pm$ 0.00 &17.17 $\pm$ 0.01 &16.40 $\pm$ 0.01 &1.19 $\pm$ 0.02 \\
         LastEpoch & 7.51 $\pm$ 0.44 &16.70 $\pm$ 2.10 &48.38 $\pm$ 17.82 &69.27 $\pm$ 3.69 &73.50 $\pm$ 4.69 &25.91 $\pm$ 5.11 &20.74 $\pm$ 3.23 &39.09 $\pm$ 11.48\\
         BestBLEU  & 6.90 $\pm$ 0.39 &14.96 $\pm$ 1.77 &46.63 $\pm$ 16.93 &68.66 $\pm$ 3.50 &72.97 $\pm$ 4.50 &24.33 $\pm$ 4.64 &19.97 $\pm$ 3.01 &35.66 $\pm$ 9.95\\
         \hline
         \rowcolor{LightBlue}\multicolumn{9}{c}{LSTM Seq2Seq } \\
         \hline
         Untrained &3.21 $\pm$ 0.00 &0.00 $\pm$ 0.00 &36.49 $\pm$ 0.00 &76.95 $\pm$ 0.01 &85.23 $\pm$ 0.01 &19.70 $\pm$ 0.01 &17.61 $\pm$ 0.00 &15.55 $\pm$ 0.01\\
         LastEpoch  & 13.48 $\pm$ 0.00 &24.85 $\pm$ 0.02 &74.23 $\pm$ 0.00 &77.77 $\pm$ 0.00 &83.73 $\pm$ 0.01 &43.32 $\pm$ 0.00 &31.48 $\pm$ 0.01 &59.64 $\pm$ 0.02 \\
         BestBLEU  &12.76 $\pm$ 0.01 &26.94 $\pm$ 0.04 &75.03 $\pm$ 0.03 &78.16 $\pm$ 0.00 &83.90 $\pm$ 0.00 &43.92 $\pm$ 0.01 &31.96 $\pm$ 0.01 &61.13 $\pm$ 0.00\\
         \hline
         \rowcolor{LightBlue}\multicolumn{9}{c}{Bi-LSTM Seq2Seq + Attention} \\
         \hline
         Untrained & 5.24 $\pm$ 0.00 &0.20 $\pm$ 0.00 &43.38 $\pm$ 0.00 &76.17 $\pm$ 0.01 &85.00 $\pm$ 0.00 &19.61 $\pm$ 0.02 &17.50 $\pm$ 0.01 &14.10 $\pm$ 0.01  \\
         LastEpoch & 14.76 $\pm$ 0.00 &23.79 $\pm$ 0.10 &75.94 $\pm$ 0.02 &79.09 $\pm$ 0.00 &85.57 $\pm$ 0.01 &40.96 $\pm$ 0.01 &28.39 $\pm$ 0.04 &55.52 $\pm$ 0.00 \\
         BestBLEU & 15.13 $\pm$ 0.01 &25.87 $\pm$ 0.05 &78.11 $\pm$ 0.02 &80.43 $\pm$ 0.02 &86.20 $\pm$ 0.00 &40.82 $\pm$ 0.01 &29.91 $\pm$ 0.02 &57.76 $\pm$ 0.00 \\
         \hline
         \rowcolor{LightBlue}\multicolumn{9}{c}{Transformer Architecture} \\
         \hline
         Untrained  & 46.64 $\pm$ 0.00 &23.15 $\pm$ 0.11 &83.55 $\pm$ 0.00 &78.12 $\pm$ 0.01 &83.93 $\pm$ 0.01 &30.92 $\pm$ 0.04 &19.59 $\pm$ 0.01 &36.67 $\pm$ 0.02 \\
         LastEpoch  & 9.46 $\pm$ 0.04 &8.45 $\pm$ 0.02 &61.31 $\pm$ 0.35 &76.18 $\pm$ 0.00 &84.90 $\pm$ 0.03 &20.27 $\pm$ 0.01 &19.08 $\pm$ 0.00 &12.41 $\pm$ 0.14 \\
         BestBLEU  & 10.43 $\pm$ 0.14 &9.71 $\pm$ 0.00 &64.42 $\pm$ 0.88 &76.10 $\pm$ 0.07 &84.20 $\pm$ 0.01 &19.83 $\pm$ 0.00 &18.34 $\pm$ 0.03 &15.35 $\pm$ 0.54\\
         \hline
    \end{tabular}
    \caption{The performance of different generative dialogue models on probe tasks in MultiWoZ dialogue data set. The Transformer model's performance decreased from initial to last epoch in majority of the tasks while SEQ2SEQ models have a learning curve.}
    \label{tab:multiwoz-probe-tasks}
\end{table*}

\begin{figure*}[h!]
\centering
\subfigure[Seq2Seq Model after 0 Epoch]{
    \includegraphics[width=0.8\columnwidth]{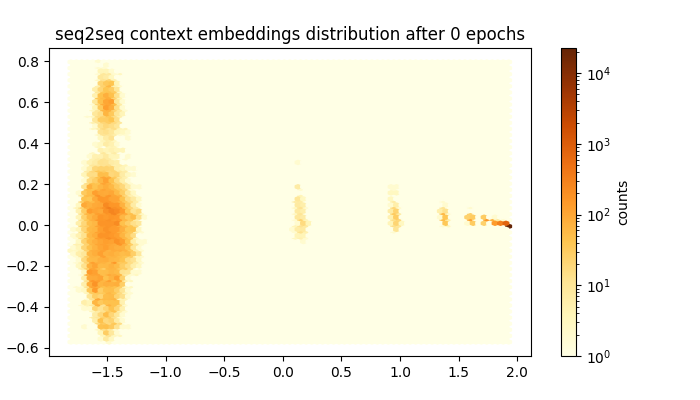}
}
\subfigure[Seq2Seq Model after Last Epoch]{
    \includegraphics[width=0.8\columnwidth]{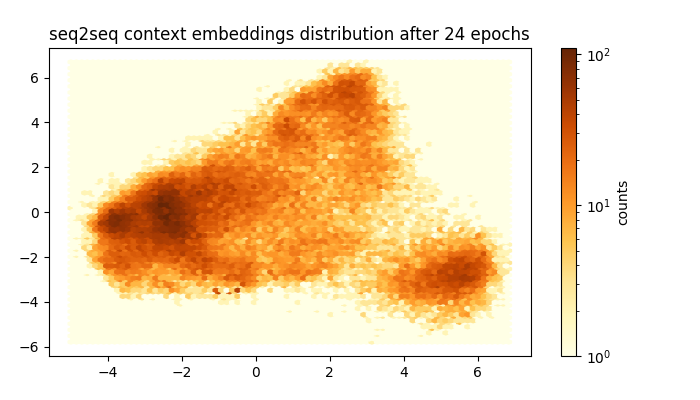}
}
\\
\subfigure[Seq2Seq Attention Model after 0 Epoch]{
    \includegraphics[width=0.8\columnwidth]{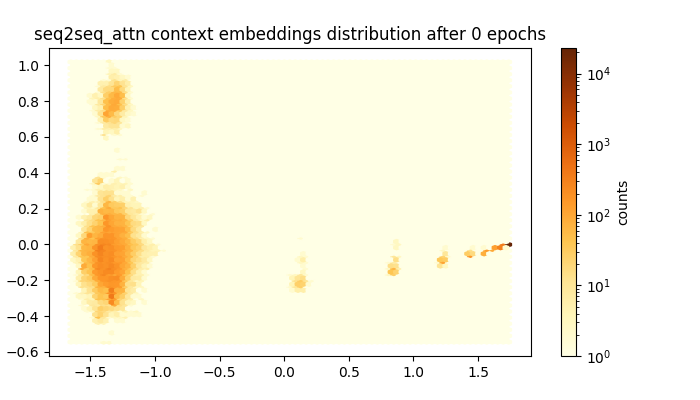}
}
\subfigure[Seq2Seq Attention Model after Last Epoch]{
    \includegraphics[width=0.8\columnwidth]{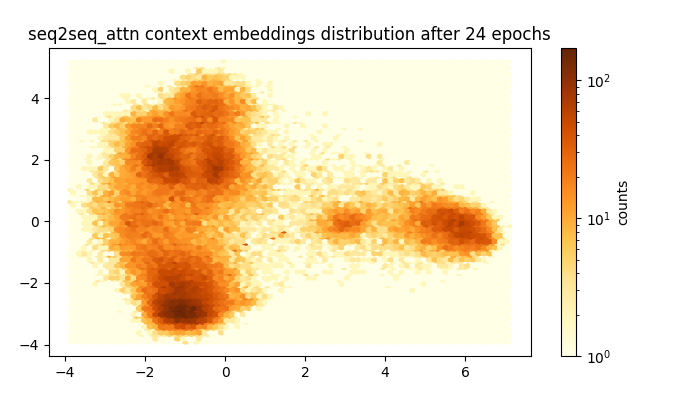}
}
\\
\subfigure[Bi-LSTM Attention Model after 0 Epoch]{
    \includegraphics[width=0.8\columnwidth]{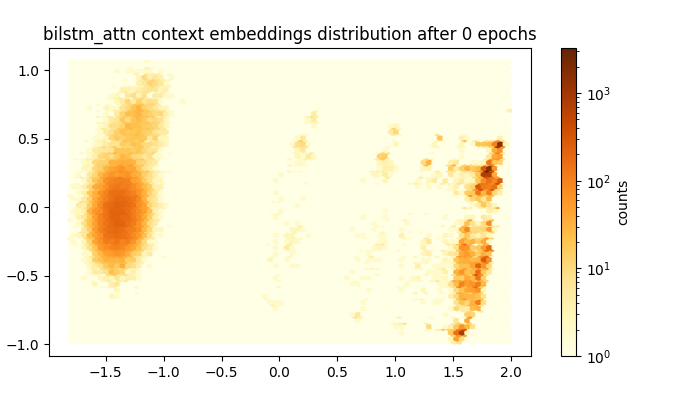}
}
\subfigure[Bi-LSTM Attention Model after Last Epoch]{
    \includegraphics[width=0.8\columnwidth]{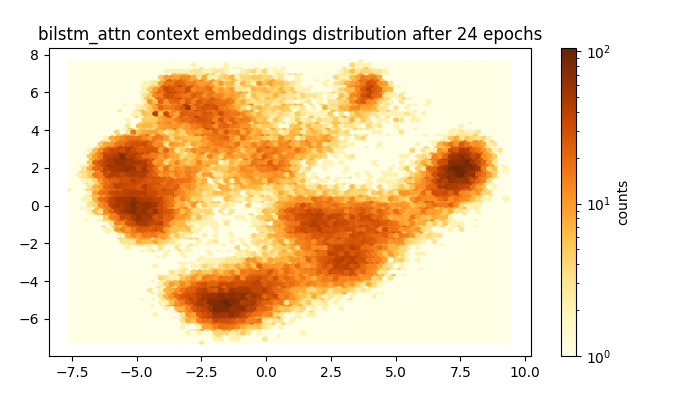}
}
\\
\subfigure[HRED Model after 0 Epoch]{
    \includegraphics[width=0.8\columnwidth]{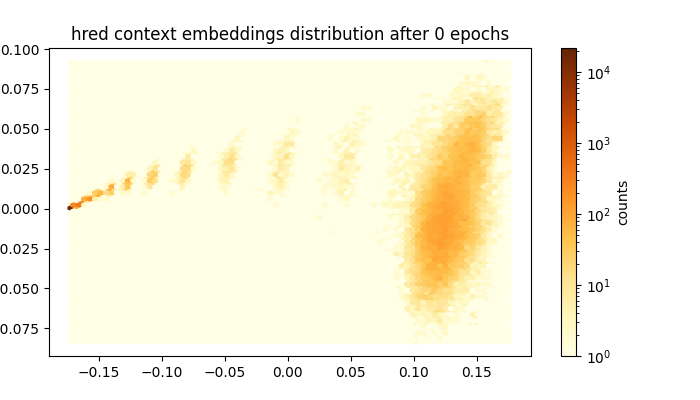}
}
\subfigure[HRED Model after Last Epoch]{
    \includegraphics[width=0.8\columnwidth]{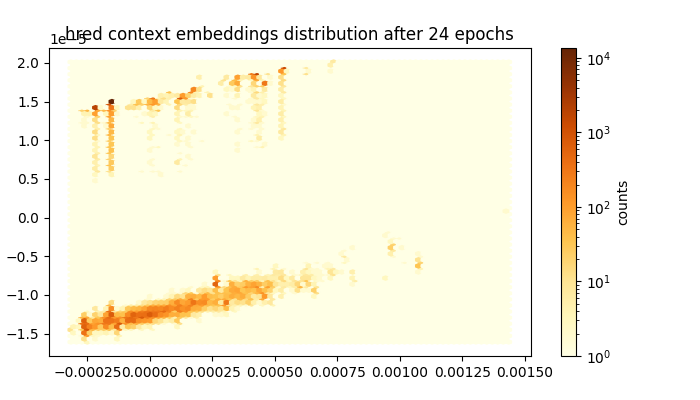}
}
\\
\subfigure[Transformer Model after 0 Epoch]{
    \includegraphics[width=0.8\columnwidth]{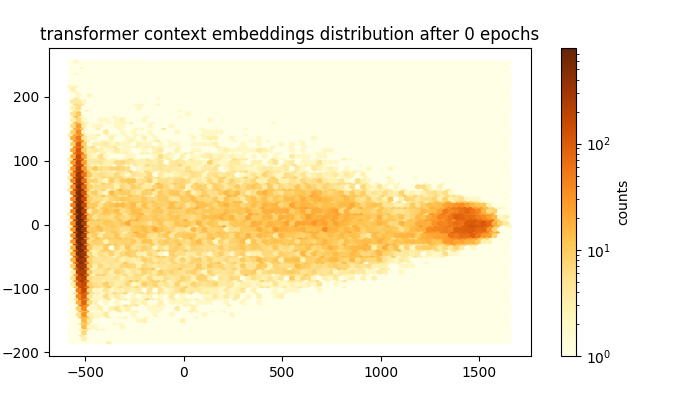}
}
\subfigure[Transformer Model after Last Epoch]{
    \includegraphics[width=0.8\columnwidth]{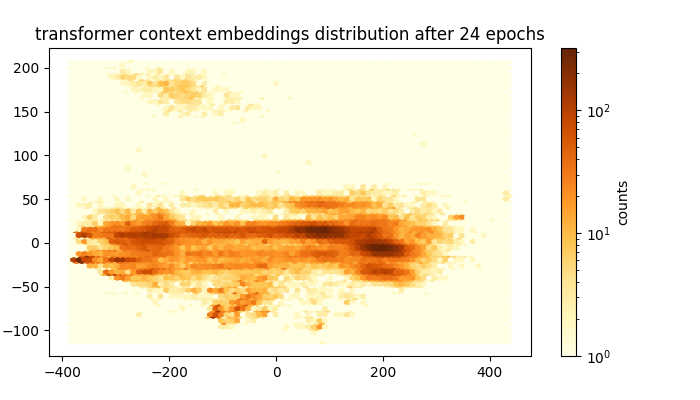}
}
\caption{Downsampled encoder hidden states on MultiWoZ dataset with PCA show that Transformer model has high capacity to encode a large dataset unlike the SEQ2SEQ models. Whereas SEQ2SEQ models improve the representation of their natural language understanding in the lower dimensional manifold as measured in the probe tasks.}
\label{fig:multi-woz-context}
\end{figure*}

To understand this phenomenon better, we downsampled the encoder representation of the contexts with PCA to 2 components (Figure \ref{fig:multi-woz-context}). Although the visualization is not an accurate indicator of what happens in the high-dimensional space, this helps in getting a reasonable understanding of the models' internals. First, the models definitely learn to cluster in the encoder hidden state that help the decoder in generating appropriate responses. Second, the range of the two axes are different for SEQ2SEQ and Transformer models. We observed that the SEQ2SEQ models, mostly, has spread out to a larger manifold from the beginning of the training to end. But, the spreading out has been constrained by the non-linear operations like \texttt{tanh} and \texttt{sigmoid}. Whereas, in the case of Transformer the manifold in an untrained model is much larger ($\sim$ 100$\times$) and eventually shrinks it during training, but still way larger ($\sim$ 40$\times$) than the embedding manifolds of SEQ2SEQ. This could be explained by the absence of \texttt{tanh} or \texttt{sigmoid} non-linearity and the stacking attention operations that only linearly combines the previous layer.

This observation reasons the need for deeper layers in Transformer when training on large data for improved performance; the attention layers help in spreading the data in a large manifold thereby the model can retain almost all of the information it was trained on. But, the reverse of generalizing from a small data is hard to come by because the model does not have sufficient direct information to cluster except the surface level signal of predicting the right tokens. This helps the Transformer model to perform well on the token prediction task or language modelling, while abstracting information and generalizing appears to be a difficult task as is observed from its performance on probing tasks.

The SEQ2SEQ models have inductive biases to squish the input through \texttt{tanh} or \texttt{sigmoid} operations. From the visualizations and from other results, we hypothesize that this aids the model in learning a regularized representation in a low-data set up. But, this can potentially be unhelpful when the input is a large set of samples and has rich structure as that requires a model to aggressively spread out. Transformer architecture can thrive in such a set up and that can be validated by the performance of large Transformer models like GPT \protect\cite{GPT}, GPT-2 \protect\cite{GPT2}, GPT-3 \protect\cite{GPT3}, BERT \protect\cite{devlin2018bert,devlin2019bert}, RoBERTa \protect\cite{liu2019roberta} etc., whereas the SEQ2SEQ models are adept at learning unsupervised structures for better understanding of the input as evaluated through the probe tasks. Also we note that the performance in probe tasks can be a pseudo metric to measure the capacity of the model in generalizing to unobserved structures in inputs in a low data scenario.

\section{Discussion}
\label{sec:discussion}
We articulated so far that mere token-level evaluation of complex language understanding tasks have high bias in automatic metrics and high variance in human evaluation. The inductive biases of models allow for different representations of the input as observed through the visualization of the embedding in Figure \ref{fig:multi-woz-context}. Although the notion of a well-structured representation is subjective, the representations can be quantitatively evaluated with the probing tasks. Such an evaluation also allows  an interpretable way for evaluating the understanding of language generation models in dialogue tasks. 

But, while deciding on the probe tasks to evaluate the models, we observed that most of the data collected for end-to-end dialogue generation tasks do not provide tasks for sanity check. Absence of probe tasks lead to draw imperfect correlations like the one between token-level accuracy and language understanding. The probe tasks show that the two are very different. To build a holistic model that can understand and perform token-level generation one may have to chose an appropriate inductive bias to train the model. At this point one may wonder, why not train the model with all the probe-tasks as auxiliary tasks for an improved performance ? Although it is a possibility, such a set up does not evaluate a model's ability to generalize to unseen dialogue tasks. Such systematic generalizations depend on the ability of a model in having a profound understanding of its input. One could potentially train a model with a fraction of the probe-tasks as auxiliary and evaluate on the rest, we leave that for future work.
\begin{table*}[h!]
    \centering
    \begin{tabular}{p{2.5cm}|c|c|c}
    \hline
         {\bf Model}& {\bf Easy} & {\bf Medium} & {\bf Hard}  \\
    \hline
        \rowcolor{Gray}{\it LSTM + Attn} & 77.6 $\pm$ 6.2 & 65.7 $\pm$ 7.6 & 44.4 $\pm$ 23.7 \\
        {\it HRED} & 72.1 $\pm$ 2.7 & 39.3 $\pm$ 5.1 & 25.4 $\pm$ 13.6 \\
        \rowcolor{Gray}{\it LSTM Seq2Seq} & 77.2 $\pm$ 5.3 & 65.7 $\pm$ 7.6 & 44.9 $\pm$ 23.5\\
        {\it BiLSTM + Attn} & 78.5 $\pm$ 6.2 & 65.6 $\pm$ 8.7 & 44.2 $\pm$ 23.3 \\
        \rowcolor{Gray}{\it Transformer} & 77.2 $\pm$ 4.9 & 43.3 $\pm$ 14.7 & 24.4 $\pm$ 16.4\\
    \hline
    \end{tabular}
    \caption{Aggregate scores of the models on performance in probe tasks of varied difficulty on MultiWoZ dataset (Higher the better). The results show that the SEQ2SEQ models perform relatively better on the medium and hard tasks while all the models perform equally good on the easy tasks.}
    \label{tab:difficulty-graded}
\end{table*}

{\textsc{Dialogue Models}} In open domain dialogues, without sufficient information to track and probe, the dialogue models cease to be a dialogue modeling tool without ways to interpret the models' language understanding. As an alternate to token-level evaluation, comparison of different model architectures can be meaningfully made with an aggregate metric on the probe tasks in three groups of difficulty -- \emph{easy} ((Ave. SEQ2SEQ) Untrained F1 $>$ .50), \emph{medium}(0.25 $<$ Untrained F1 $\geq$ 0.50), and \emph{hard} (Untrained F1 $<$ 0.25). Such an analysis, as shown in Table \ref{tab:difficulty-graded}, allows better inspection of the model results and a fairer comparison between the models. We can see from Table \ref{tab:difficulty-graded} that the models have difficulty in solving harder probe tasks. The results can be used as motivation for coming up with novel inductive biases for neural architectures that address one or a group of aspects in the language understanding of generative dialogue models. 

{\textsc{Dialogue Datasets}} The challenges in dialogue modeling has been evolving majorly because of the complex datasets. But, datasets on chit-chat dialogues often have little to no auxiliary tasks to evaluate the dialogue management abilities of a model. This limits the practitioners to validate the models only on the text generation abilities which, in this paper, is shown to have no correlation with the model's ability to manage internal states in a dialogue. 

Goal oriented datasets provide rich set of information that allows evaluating the internal workings of generative dialogue models. Further, probe tasks can also serve as a way to compare datasets in similar domain to rank them on the difficulty in comprehending the input context. Such a set up invites datasets that challenge models on their understanding. We observed that the chit-chat dialogue datasets need to have a richer set of probing tasks; and dialogue state tracking for non-goal oriented dialogues could be a way forward.

\section{Conclusion}
We propose a set of probing tasks to compare end-to-end generative dialogue models. We observed that mimicking surface level token prediction is easier for models than learning representations that understand the context. The results of the experiments on probe tasks showed that SEQ2SEQ models perform better than transformer model in encoding information in the context that often goes unsupervised. We also found some probe tasks that all of the models find difficult to solve; this invites novel architectures that can handle the language understanding aspects in dialogue generation. Although language modeling is required for a dialogue model, the performance in token prediction alone cannot be a proxy for the model's ability to understand a conversation. Hence, systematically identifying issues with probe tasks can help in building better models for the task and collecting datasets that allow holistic evaluation of a model's performance. 
\section*{Acknowledgements}
We thank Janardhanan Rajendran (University of Michigan) and Koustuv Sinha (McGill University) for their discussions. We gratefully acknowledge financial support by the Natural Sciences and Engineering Research Council of Canada (NSERC) and Canada CIFAR AI Chair. We also thank McGill University for the computing resources which were immensely helpful to carry out this research. Sarath Chandar is supported by a Canada CIFAR AI Chair and an NSERC Discovery Grant.


\bibliography{emnlp2020}

\begin{thebibliography}{73}
\expandafter\ifx\csname natexlab\endcsname\relax\def\natexlab#1{#1}\fi

\bibitem[{Anand et~al.(2019)Anand, Racah, Ozair, Bengio, C{\^o}t{\'e}, and
  Hjelm}]{anand2019unsupervised}
Ankesh Anand, Evan Racah, Sherjil Ozair, Yoshua Bengio, Marc-Alexandre
  C{\^o}t{\'e}, and R~Devon Hjelm. 2019.
\newblock Unsupervised state representation learning in atari.
\newblock In \emph{NeurIPS}.

\bibitem[{Asri et~al.(2017)Asri, Schulz, Sharma, Zumer, Harris, Fine, Mehrotra,
  and Suleman}]{asri2017frames}
Layla~El Asri, Hannes Schulz, Shikhar Sharma, Jeremie Zumer, Justin Harris,
  Emery Fine, Rahul Mehrotra, and Kaheer Suleman. 2017.
\newblock Frames: A corpus for adding memory to goal-oriented dialogue systems.
\newblock \emph{arXiv}.

\bibitem[{Bahdanau et~al.(2014)Bahdanau, Cho, and Bengio}]{nmtbahadanau}
Dzmitry Bahdanau, Kyunghyun Cho, and Yoshua Bengio. 2014.
\newblock Neural machine translation by jointly learning to align and
  translate.
\newblock In \emph{arXiv}.

\bibitem[{Belinkov and Glass(2019)}]{belinkov2019analysis}
Yonatan Belinkov and James Glass. 2019.
\newblock Analysis methods in neural language processing: A survey.
\newblock \emph{TACL}.

\bibitem[{Bengio et~al.(2003)Bengio, Ducharme, Vincent, and
  Jauvin}]{bengio2003neural}
Yoshua Bengio, R{\'e}jean Ducharme, Pascal Vincent, and Christian Jauvin. 2003.
\newblock A neural probabilistic language model.
\newblock In \emph{Journal of machine learning research}.

\bibitem[{Bordes et~al.(2016)Bordes, Boureau, and Weston}]{bordes2016learning}
Antoine Bordes, Y-Lan Boureau, and Jason Weston. 2016.
\newblock Learning end-to-end goal-oriented dialog.
\newblock In \emph{arXiv}.

\bibitem[{Brown et~al.(2020)Brown, Mann, Ryder, Subbiah, Kaplan, Dhariwal,
  Neelakantan, Shyam, Sastry, Askell et~al.}]{GPT3}
Tom~B Brown, Benjamin Mann, Nick Ryder, Melanie Subbiah, Jared Kaplan, Prafulla
  Dhariwal, Arvind Neelakantan, Pranav Shyam, Girish Sastry, Amanda Askell,
  et~al. 2020.
\newblock Language models are few-shot learners.
\newblock \emph{arXiv}.

\bibitem[{Budzianowski et~al.(2018)Budzianowski, Wen, Tseng, Casanueva, Stefan,
  Osman, and Ga{\v{s}}i\'c}]{budzianowski2018multiwoz}
Pawe{\l} Budzianowski, Tsung-Hsien Wen, Bo-Hsiang Tseng, I{\~n}igo Casanueva,
  Ultes Stefan, Ramadan Osman, and Milica Ga{\v{s}}i\'c. 2018.
\newblock Multiwoz - a large-scale multi-domain wizard-of-oz dataset for
  task-oriented dialogue modelling.
\newblock In \emph{Proceedings of EMNLP}.

\bibitem[{Buhrmester et~al.(2016)Buhrmester, Kwang, and
  Gosling}]{buhrmester2016amazon}
Michael Buhrmester, Tracy Kwang, and Samuel~D Gosling. 2016.
\newblock Amazon's mechanical turk: A new source of inexpensive, yet
  high-quality data?
\newblock \emph{American Psychological Association}.

\bibitem[{Carlson et~al.(2010)Carlson, Betteridge, Kisiel, and
  Settles}]{carlson2010toward}
Andrew Carlson, Justin Betteridge, Bryan Kisiel, and Burr Settles. 2010.
\newblock Toward an architecture for never-ending language learning.
\newblock In \emph{Proceedings of AAAI}.

\bibitem[{Conneau et~al.(2018)Conneau, Kruszewski, Lample, Barrault, and
  Baroni}]{conneau2018you}
Alexis Conneau, Germ{\'a}n Kruszewski, Guillaume Lample, Lo{\"\i}c Barrault,
  and Marco Baroni. 2018.
\newblock What you can cram into a single \$$\&!$\#* vector: Probing sentence
  embeddings for linguistic properties.
\newblock In \emph{Proceedings of ACL}.

\bibitem[{Dave et~al.(2003)Dave, Lawrence, and Pennock}]{dave2003mining}
Kushal Dave, Steve Lawrence, and David~M Pennock. 2003.
\newblock Mining the peanut gallery: Opinion extraction and semantic
  classification of product reviews.
\newblock In \emph{Proceedings of the 12th international conference on World
  Wide Web}.

\bibitem[{Devlin et~al.(2018)Devlin, Chang, Lee, and
  Toutanova}]{devlin2018bert}
Jacob Devlin, Ming-Wei Chang, Kenton Lee, and Kristina Toutanova. 2018.
\newblock Bert: Pre-training of deep bidirectional transformers for language
  understanding.
\newblock \emph{arXiv}.

\bibitem[{Devlin et~al.(2019)Devlin, Chang, Lee, and
  Toutanova}]{devlin2019bert}
Jacob Devlin, Ming-Wei Chang, Kenton Lee, and Kristina Toutanova. 2019.
\newblock Bert: Pre-training of deep bidirectional transformers for language
  understanding.
\newblock In \emph{Proceedings of the 2019 Conference of the North American
  Chapter of the Association for Computational Linguistics (NAACL)}.

\bibitem[{Dinan et~al.(2018)Dinan, Roller, Shuster, Fan, Auli, and
  Weston}]{dinan2018wizard}
Emily Dinan, Stephen Roller, Kurt Shuster, Angela Fan, Michael Auli, and Jason
  Weston. 2018.
\newblock Wizard of wikipedia: Knowledge-powered conversational agents.
\newblock \emph{arXiv}.

\bibitem[{Donahue et~al.(2015)Donahue, Anne~Hendricks, Guadarrama, Rohrbach,
  Venugopalan, Saenko, and Darrell}]{donahue2015long}
Jeffrey Donahue, Lisa Anne~Hendricks, Sergio Guadarrama, Marcus Rohrbach,
  Subhashini Venugopalan, Kate Saenko, and Trevor Darrell. 2015.
\newblock Long-term recurrent convolutional networks for visual recognition and
  description.
\newblock In \emph{CVPR}.

\bibitem[{Elazar et~al.(2020)Elazar, Ravfogel, Jacovi, and
  Goldberg}]{elazar2020bert}
Yanai Elazar, Shauli Ravfogel, Alon Jacovi, and Yoav Goldberg. 2020.
\newblock When bert forgets how to pos: Amnesic probing of linguistic
  properties and mlm predictions.
\newblock \emph{arXiv}.

\bibitem[{Gambhir and Gupta(2017)}]{gambhir2017recent}
Mahak Gambhir and Vishal Gupta. 2017.
\newblock Recent automatic text summarization techniques: a survey.
\newblock \emph{Artificial Intelligence Review}.

\bibitem[{Guo et~al.(2018)Guo, Tang, Duan, Zhou, and Yin}]{guo2018dialog}
Daya Guo, Duyu Tang, Nan Duan, Ming Zhou, and Jian Yin. 2018.
\newblock Dialog-to-action: Conversational question answering over a
  large-scale knowledge base.
\newblock In \emph{Advances in Neural Information Processing Systems}.

\bibitem[{Henderson et~al.(2014)Henderson, Thomson, and
  Williams}]{henderson2014second}
Matthew Henderson, Blaise Thomson, and Jason~D Williams. 2014.
\newblock The second dialog state tracking challenge.
\newblock In \emph{SIGDIAL}.

\bibitem[{Jawahar et~al.(2019)Jawahar, Sagot, and Seddah}]{jawahar2019does}
Ganesh Jawahar, Beno{\^\i}t Sagot, and Djam{\'e} Seddah. 2019.
\newblock What does bert learn about the structure of language?
\newblock In \emph{ACL}.

\bibitem[{Kahneman(2014)}]{kahneman2014thinking}
Daniel Kahneman. 2014.
\newblock Thinking, fast and slow.
\newblock \emph{Journal of Managemet Research in Emerging Economies}.

\bibitem[{Kalchbrenner and Blunsom(2013)}]{kalchbrenner-blunsom-2013-recurrent}
Nal Kalchbrenner and Phil Blunsom. 2013.
\newblock Recurrent continuous translation models.
\newblock In \emph{Proceedings of EMNLP}.

\bibitem[{Kingma and Ba(2014)}]{kingma2014adam}
Diederik~P Kingma and Jimmy Ba. 2014.
\newblock Adam: A method for stochastic optimization.
\newblock \emph{arXiv}.

\bibitem[{Klein and Manning(2003)}]{klein2003accurate}
Dan Klein and Christopher~D Manning. 2003.
\newblock Accurate unlexicalized parsing.
\newblock In \emph{ACL}.

\bibitem[{Koomen and Pol(1999)}]{koomen1999test}
Tim Koomen and Martin Pol. 1999.
\newblock \emph{Test process improvement: a practical step-by-step guide to
  structured testing}.
\newblock Addison-Wesley Longman Publishing Co., Inc.

\bibitem[{Kupiec et~al.(1995)Kupiec, Pedersen, and Chen}]{kupiec1995trainable}
Julian Kupiec, Jan Pedersen, and Francine Chen. 1995.
\newblock A trainable document summarizer.
\newblock In \emph{Proceedings of the 18th annual international ACM SIGIR
  conference on Research and development in information retrieval}.

\bibitem[{Landauer and Dumais(1997)}]{landauer1997solution}
Thomas~K Landauer and Susan~T Dumais. 1997.
\newblock A solution to plato's problem: The latent semantic analysis theory of
  acquisition, induction, and representation of knowledge.
\newblock \emph{Psychological review}.

\bibitem[{Lang et~al.(1990)Lang, Waibel, and Hinton}]{rnnlm}
Kevin~J Lang, Alex~H Waibel, and Geoffrey~E Hinton. 1990.
\newblock A time-delay neural network architecture for isolated word
  recognition.
\newblock In \emph{Neural networks}.

\bibitem[{Lavie and Agarwal(2007)}]{lavie2007meteor}
Alon Lavie and Abhaya Agarwal. 2007.
\newblock Meteor: An automatic metric for mt evaluation with high levels of
  correlation with human judgments.
\newblock In \emph{Proceedings of the Second Workshop on Statistical Machine
  Translation}. Association for Computational Linguistics.

\bibitem[{LeCun et~al.(1998)LeCun, Bottou, Bengio, and
  Haffner}]{lecun1998gradient}
Yann LeCun, L{\'e}on Bottou, Yoshua Bengio, and Patrick Haffner. 1998.
\newblock Gradient-based learning applied to document recognition.
\newblock \emph{Proceedings of the IEEE}.

\bibitem[{Li et~al.(2015)Li, Galley, Brockett, Gao, and
  Dolan}]{li2015diversity}
Jiwei Li, Michel Galley, Chris Brockett, Jianfeng Gao, and Bill Dolan. 2015.
\newblock A diversity-promoting objective function for neural conversation
  models.
\newblock \emph{arXiv}.

\bibitem[{Li et~al.(2016)Li, Monroe, Ritter, and Jurafsky}]{li2016deep}
Jiwei Li, Will Monroe, Alan Ritter, and Dan Jurafsky. 2016.
\newblock Deep reinforcement learning for dialogue generation.
\newblock \emph{Proceedings of EMNLP}.

\bibitem[{Li et~al.(2017)Li, Monroe, Shi, Ritter, and
  Jurafsky}]{li2017adversarial}
Jiwei Li, Will Monroe, Tianlin Shi, Alan Ritter, and Dan Jurafsky. 2017.
\newblock Adversarial learning for neural dialogue generation.
\newblock In \emph{arXiv}.

\bibitem[{Li et~al.(2019)Li, Weston, and Roller}]{li2019acute}
Margaret Li, Jason Weston, and Stephen Roller. 2019.
\newblock Acute-eval: Improved dialogue evaluation with optimized questions and
  multi-turn comparisons.
\newblock \emph{arXiv}.

\bibitem[{Lin(2004)}]{lin2004rouge}
Chin-Yew Lin. 2004.
\newblock Rouge: A package for automatic evaluation of summaries.
\newblock In \emph{Text summarization branches out}.

\bibitem[{Liu et~al.(2016)Liu, Lowe, Serban, Noseworthy, Charlin, and
  Pineau}]{liu2016not}
Chia-Wei Liu, Ryan Lowe, Iulian~Vlad Serban, Mike Noseworthy, Laurent Charlin,
  and Joelle Pineau. 2016.
\newblock How not to evaluate your dialogue system: An empirical study of
  unsupervised evaluation metrics for dialogue response generation.
\newblock In \emph{Proceedings of EMNLP}.

\bibitem[{Liu et~al.(2019)Liu, Ott, Goyal, Du, Joshi, Chen, Levy, Lewis,
  Zettlemoyer, and Stoyanov}]{liu2019roberta}
Yinhan Liu, Myle Ott, Naman Goyal, Jingfei Du, Mandar Joshi, Danqi Chen, Omer
  Levy, Mike Lewis, Luke Zettlemoyer, and Veselin Stoyanov. 2019.
\newblock Roberta: A robustly optimized bert pretraining approach.
\newblock \emph{arXiv}.

\bibitem[{Lowe et~al.(2017)Lowe, Noseworthy, Serban, Angelard-Gontier, Bengio,
  and Pineau}]{lowe2017towards}
Ryan Lowe, Michael Noseworthy, Iulian~V Serban, Nicolas Angelard-Gontier,
  Yoshua Bengio, and Joelle Pineau. 2017.
\newblock Towards an automatic turing test: Learning to evaluate dialogue
  responses.
\newblock \emph{arXiv}.

\bibitem[{Lowe et~al.(2015{\natexlab{a}})Lowe, Pow, Serban, Charlin, and
  Pineau}]{lowe2015incorporating}
Ryan Lowe, Nissan Pow, Iulian Serban, Laurent Charlin, and Joelle Pineau.
  2015{\natexlab{a}}.
\newblock Incorporating unstructured textual knowledge sources into neural
  dialogue systems.
\newblock In \emph{In NeurIPS Workshop on Machine Learning for Spoken Language
  Understanding}.

\bibitem[{Lowe et~al.(2015{\natexlab{b}})Lowe, Pow, Serban, and
  Pineau}]{lowe2015ubuntu}
Ryan Lowe, Nissan Pow, Iulian Serban, and Joelle Pineau. 2015{\natexlab{b}}.
\newblock The ubuntu dialogue corpus: A large dataset for research in
  unstructured multi-turn dialogue systems.
\newblock In \emph{arXiv}.

\bibitem[{Luhn(1958)}]{luhn1958automatic}
Hans~Peter Luhn. 1958.
\newblock The automatic creation of literature abstracts.
\newblock \emph{IBM Journal of research and development}.

\bibitem[{Miller et~al.(2017)Miller, Feng, Fisch, Lu, Batra, Bordes, Parikh,
  and Weston}]{miller2017parlai}
Alexander~H Miller, Will Feng, Adam Fisch, Jiasen Lu, Dhruv Batra, Antoine
  Bordes, Devi Parikh, and Jason Weston. 2017.
\newblock Parlai: A dialog research software platform.
\newblock \emph{arXiv}.

\bibitem[{Neelakantan et~al.(2019)Neelakantan, Yavuz, Narang, Prasad, Goodrich,
  Duckworth, Sankar, and Yan}]{neelakantan2019neural}
Arvind Neelakantan, Semih Yavuz, Sharan Narang, Vishaal Prasad, Ben Goodrich,
  Daniel Duckworth, Chinnadhurai Sankar, and Xifeng Yan. 2019.
\newblock Neural assistant: Joint action prediction, response generation, and
  latent knowledge reasoning.
\newblock \emph{arXiv}.

\bibitem[{Papineni et~al.(2002)Papineni, Roukos, Ward, and
  Zhu}]{papineni2002bleu}
Kishore Papineni, Salim Roukos, Todd Ward, and Wei-Jing Zhu. 2002.
\newblock Bleu: a method for automatic evaluation of machine translation.
\newblock In \emph{Proceedings of the 40th annual meeting on association for
  computational linguistics}. Association for Computational Linguistics.

\bibitem[{Parthasarathi and Pineau(2018)}]{parthasarathi2018extending}
Prasanna Parthasarathi and Joelle Pineau. 2018.
\newblock Extending neural generative conversational model using external
  knowledge sources.
\newblock In \emph{Proceedings of EMNLP}.

\bibitem[{Paszke et~al.(2017)Paszke, Gross, Chintala, Chanan, Yang, DeVito,
  Lin, Desmaison, Antiga, and Lerer}]{paszke2017pytorch}
Adam Paszke, Sam Gross, Soumith Chintala, Gregory Chanan, Edward Yang, Zachary
  DeVito, Zeming Lin, Alban Desmaison, Luca Antiga, and Adam Lerer. 2017.
\newblock Automatic differentiation in pytorch.
\newblock \emph{NIPS-W}.

\bibitem[{Pedregosa et~al.(2011)Pedregosa, Varoquaux, Gramfort, Michel,
  Thirion, Grisel, Blondel, Prettenhofer, Weiss, Dubourg, Vanderplas, Passos,
  Cournapeau, Brucher, Perrot, and Duchesnay}]{scikit-learn}
F.~Pedregosa, G.~Varoquaux, A.~Gramfort, V.~Michel, B.~Thirion, O.~Grisel,
  M.~Blondel, P.~Prettenhofer, R.~Weiss, V.~Dubourg, J.~Vanderplas, A.~Passos,
  D.~Cournapeau, M.~Brucher, M.~Perrot, and E.~Duchesnay. 2011.
\newblock {Scikit-learn: Machine Learning in Python }.
\newblock \emph{Journal of Machine Learning Research}.

\bibitem[{Radford et~al.(2018)Radford, Narasimhan, Salimans, and
  Sutskever}]{GPT}
Alec Radford, Karthik Narasimhan, Tim Salimans, and Ilya Sutskever. 2018.
\newblock Improving language understanding by generative pre-training.

\bibitem[{Radford et~al.(2019)Radford, Wu, Child, Luan, Amodei, and
  Sutskever}]{GPT2}
Alec Radford, Jeffrey Wu, Rewon Child, David Luan, Dario Amodei, and Ilya
  Sutskever. 2019.
\newblock Language models are unsupervised multitask learners.
\newblock \emph{OpenAI Blog}.

\bibitem[{Rajpurkar et~al.(2016)Rajpurkar, Zhang, Lopyrev, and
  Liang}]{rajpurkar2016squad}
Pranav Rajpurkar, Jian Zhang, Konstantin Lopyrev, and Percy Liang. 2016.
\newblock Squad: 100,000+ questions for machine comprehension of text.
\newblock \emph{arXiv}.

\bibitem[{Ramsauer et~al.(2020)Ramsauer, Sch{\"a}fl, Lehner, Seidl, Widrich,
  Gruber, Holzleitner, Pavlovi{\'c}, Sandve, Greiff
  et~al.}]{ramsauer2020hopfield}
Hubert Ramsauer, Bernhard Sch{\"a}fl, Johannes Lehner, Philipp Seidl, Michael
  Widrich, Lukas Gruber, Markus Holzleitner, Milena Pavlovi{\'c}, Geir~Kjetil
  Sandve, Victor Greiff, et~al. 2020.
\newblock Hopfield networks is all you need.
\newblock \emph{arXiv}.

\bibitem[{Reddy et~al.(2019)Reddy, Chen, and Manning}]{reddy2019coqa}
Siva Reddy, Danqi Chen, and Christopher~D Manning. 2019.
\newblock Coqa: A conversational question answering challenge.
\newblock \emph{Transactions of the Association for Computational Linguistics}.

\bibitem[{Ritter et~al.(2011)Ritter, Cherry, and Dolan}]{ritter2011data}
Alan Ritter, Colin Cherry, and William~B Dolan. 2011.
\newblock Data-driven response generation in social media.
\newblock In \emph{Proceedings of the conference on empirical methods in
  natural language processing}.

\bibitem[{Rus and Lintean(2012)}]{rus2012optimal}
Vasile Rus and Mihai Lintean. 2012.
\newblock An optimal assessment of natural language student input using
  word-to-word similarity metrics.
\newblock In \emph{International Conference on Intelligent Tutoring Systems},
  pages 675--676. Springer.

\bibitem[{Rush(2018)}]{rush2018annotated}
Alexander~M Rush. 2018.
\newblock The annotated transformer.
\newblock In \emph{Proceedings of workshop for NLP open source software
  (NLP-OSS)}.

\bibitem[{Sankar et~al.(2019)Sankar, Subramanian, Pal, Chandar, and
  Bengio}]{chinnapaper}
Chinnadhurai Sankar, Sandeep Subramanian, Chris Pal, Sarath Chandar, and Yoshua
  Bengio. 2019.
\newblock Do neural dialog systems use the conversation history effectively? an
  empirical study.
\newblock In \emph{Association of Computational Linguistics}.

\bibitem[{Scheepers(2017)}]{scheepers2017compositionality}
Thijs Scheepers. 2017.
\newblock Improving the compositionality of word embeddings.
\newblock Master's thesis, Universiteit van Amsterdam.

\bibitem[{Schuster and Paliwal(1997)}]{bidirectional}
Mike Schuster and Kuldip~K Paliwal. 1997.
\newblock Bidirectional recurrent neural networks.
\newblock In \emph{IEEE Transactions on Signal Processing}.

\bibitem[{Serban et~al.(2015)Serban, Sordoni, Bengio, Courville, and
  Pineau}]{serban2015hierarchical}
Iulian~V Serban, Alessandro Sordoni, Yoshua Bengio, Aaron Courville, and Joelle
  Pineau. 2015.
\newblock Hierarchical neural network generative models for movie dialogues.
\newblock \emph{arXiv}.

\bibitem[{Sinha et~al.(2020)Sinha, Parthasarathi, Wang, Lowe, Hamilton, and
  Pineau}]{sinha2020learning}
Koustuv Sinha, Prasanna Parthasarathi, Jasmine Wang, Ryan Lowe, William~L
  Hamilton, and Joelle Pineau. 2020.
\newblock Learning an unreferenced metric for online dialogue evaluation.
\newblock \emph{In Proceedings of ACL}.

\bibitem[{Sordoni et~al.(2015)Sordoni, Bengio, Vahabi, Lioma, Grue~Simonsen,
  and Nie}]{sordoni2015hierarchical}
Alessandro Sordoni, Yoshua Bengio, Hossein Vahabi, Christina Lioma, Jakob
  Grue~Simonsen, and Jian-Yun Nie. 2015.
\newblock A hierarchical recurrent encoder-decoder for generative context-aware
  query suggestion.
\newblock In \emph{Proceedings of the 24th ACM International on Conference on
  Information and Knowledge Management}.

\bibitem[{Tallec et~al.(2019)Tallec, Blier, and Kalainathan}]{worldmodelblog}
Corentin Tallec, Léonard Blier, and Diviyan Kalainathan. 2019.
\newblock Reproducing "world models": Is training the recurrent network really
  needed ?
\newblock \emph{https://ctallec.github.io/world-models/}.

\bibitem[{Tao et~al.(2018)Tao, Mou, Zhao, and Yan}]{tao2018ruber}
Chongyang Tao, Lili Mou, Dongyan Zhao, and Rui Yan. 2018.
\newblock Ruber: An unsupervised method for automatic evaluation of open-domain
  dialog systems.
\newblock In \emph{Thirty-Second AAAI Conference on Artificial Intelligence}.

\bibitem[{Vaswani et~al.(2017)Vaswani, Shazeer, Parmar, Uszkoreit, Jones,
  Gomez, Kaiser, and Polosukhin}]{vaswani2017attention}
Ashish Vaswani, Noam Shazeer, Niki Parmar, Jakob Uszkoreit, Llion Jones,
  Aidan~N Gomez, {\L}ukasz Kaiser, and Illia Polosukhin. 2017.
\newblock Attention is all you need.
\newblock In \emph{Advances in neural information processing systems}.

\bibitem[{Vinyals and Le(2015)}]{vinyals2015neural}
Oriol Vinyals and Quoc Le. 2015.
\newblock A neural conversational model.
\newblock \emph{arXiv}.

\bibitem[{Vinyals et~al.(2015)Vinyals, Toshev, Bengio, and
  Erhan}]{vinyals2015show}
Oriol Vinyals, Alexander Toshev, Samy Bengio, and Dumitru Erhan. 2015.
\newblock Show and tell: A neural image caption generator.
\newblock In \emph{CVPR}.

\bibitem[{Wang and Waibel(1997)}]{wang1997decoding}
Ye-Yi Wang and Alex Waibel. 1997.
\newblock Decoding algorithm in statistical machine translation.
\newblock In \emph{European Chapter of the Association for Computational
  Linguistics}.

\bibitem[{Wieting et~al.(2015)Wieting, Bansal, Gimpel, and
  Livescu}]{wieting2015towards}
John Wieting, Mohit Bansal, Kevin Gimpel, and Karen Livescu. 2015.
\newblock Towards universal paraphrastic sentence embeddings.
\newblock \emph{arXiv}.

\bibitem[{Williams and Zipser(1989)}]{teacherforcing}
Ronald~J Williams and David Zipser. 1989.
\newblock A learning algorithm for continually running fully recurrent neural
  networks.
\newblock \emph{Neural computation}.

\bibitem[{Xu et~al.(2015)Xu, Ba, Kiros, Cho, Courville, Salakhudinov, Zemel,
  and Bengio}]{xu2015show}
Kelvin Xu, Jimmy Ba, Ryan Kiros, Kyunghyun Cho, Aaron Courville, Ruslan
  Salakhudinov, Rich Zemel, and Yoshua Bengio. 2015.
\newblock Show, attend and tell: Neural image caption generation with visual
  attention.
\newblock In \emph{ICML}.

\bibitem[{Yampolskiy(2013)}]{yampolskiy2013aicomplete}
Roman~V Yampolskiy. 2013.
\newblock Turing test as a defining feature of ai-completeness.
\newblock In \emph{Artificial intelligence, evolutionary computing and
  metaheuristics}. Springer.

\bibitem[{Zhang et~al.(2018)Zhang, Dinan, Urbanek, Szlam, Kiela, and
  Weston}]{zhang2018personalizing}
Saizheng Zhang, Emily Dinan, Jack Urbanek, Arthur Szlam, Douwe Kiela, and Jason
  Weston. 2018.
\newblock Personalizing dialogue agents: I have a dog, do you have pets too?
\newblock \emph{arXiv}.

\end{thebibliography}
\bibliographystyle{acl_natbib}

\appendix
\section*{Appendix}
\section{Model Parameters}
\label{sec:model-parameters}
\begin{itemize}
    \item For SEQ2SEQ models, we used a 256 unit hidden size LSTM with 2 layers and a 128 unit input embedding dimension. The learning rate we used for all the models is 4E-3. 
    \item For Transformer, we used a 512 unit hidden size, 512 unit input embedding dimension, 2 attention header and 4 layers. 
    \item We used Adam as the optimizer to optimize on the cross-entropy loss. 
    \item We averaged the results over 3 different seeds.
    \item We used a truncated history of last 100 tokens as context to keep the training uniform across the models.
\end{itemize}

\section{Data Preparation}
\label{sec:data-preparation}
\subsection{MultiWoZ}
We use the minimal information in the annotated json dataset to add additional details to the dialogue state. The annotation has texts of user and agent utterances with slot and value pairs parsed from the user utterance. We parse and process the slot-value pairs data into information like task topics, recent slot, among others used in the probe tasks. For UtteranceLoc, we bucket the utterances based on the position they occur in a conversation.

\subsection{PersonaChat}
PersonaChat dataset has persona details along with text utterances from two different users. We extract only the non-stop words from the persona information for PersonalInfo task. For WordCont, we order the vocab descending frequency and select 500 words that occur for 1000-3000 times in the dataset.

\section{Distribution of Information}
\label{sec:information-distribution}
\begin{figure*}[h!]
\centering
\subfigure[Distribution of Topics]{
    \includegraphics[width=0.8\columnwidth]{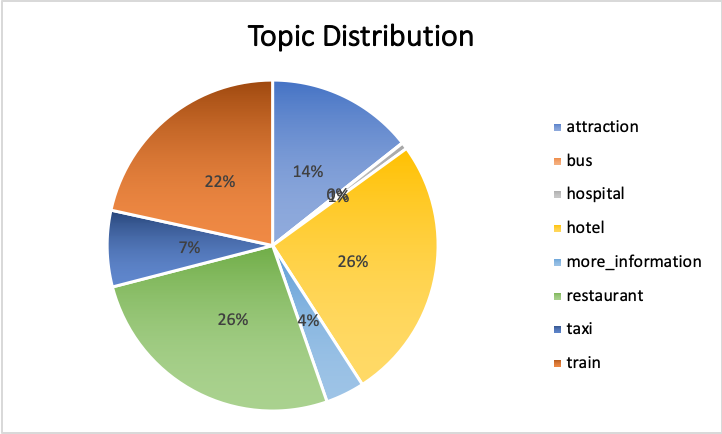}
}
\subfigure[\#Topics Distribution in a Conversation]{
    \includegraphics[width=0.8\columnwidth]{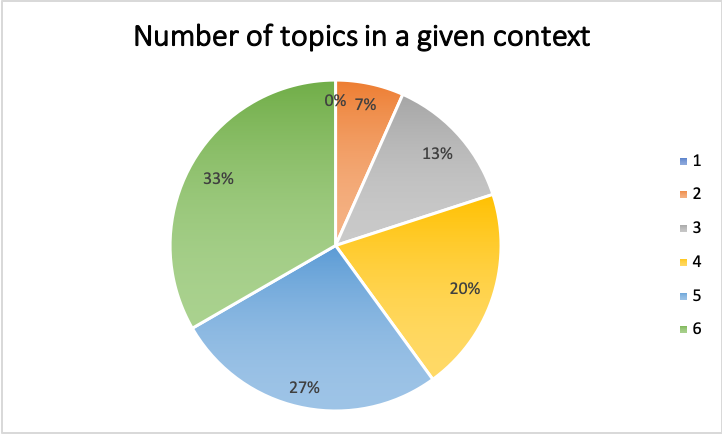}
}
\\
\subfigure[Number of Information in Single Utterance]{
    \includegraphics[width=0.8\columnwidth]{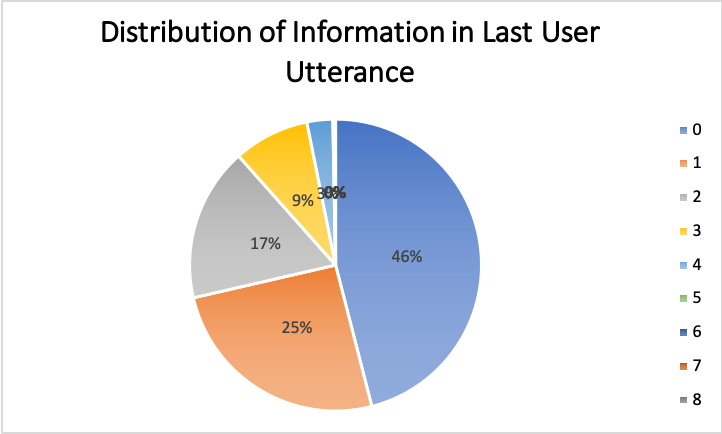}
}
\subfigure[Number of information Repeating in a Context]{
    \includegraphics[width=0.8\columnwidth]{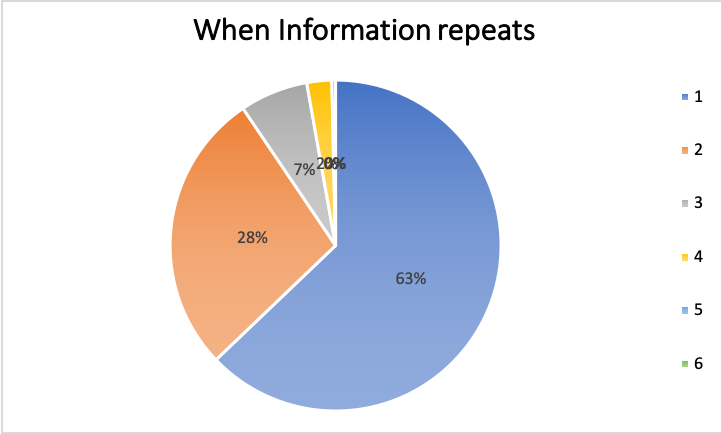}
}
\\
\subfigure[Distribution of Single and Multiple Tasks]{
    \includegraphics[width=0.8\columnwidth]{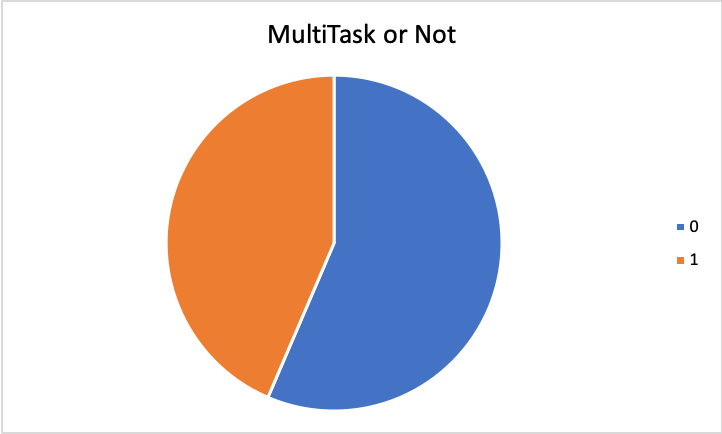}
}
\subfigure[Location of Utterance Distribution]{
    \includegraphics[width=0.8\columnwidth]{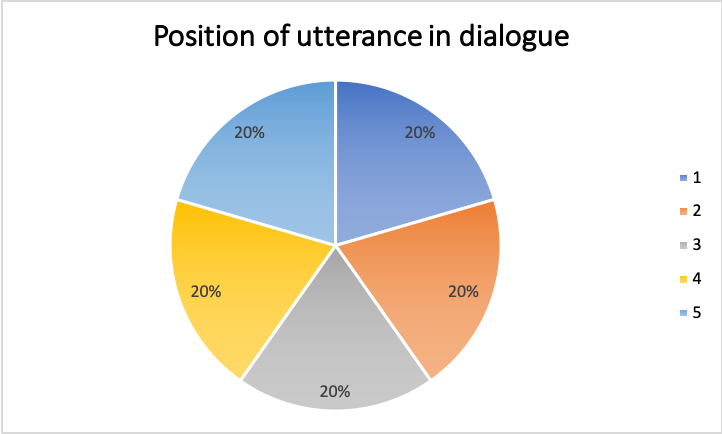}
}
\\
\subfigure[Response Length Distribution]{
    \includegraphics[width=0.8\columnwidth]{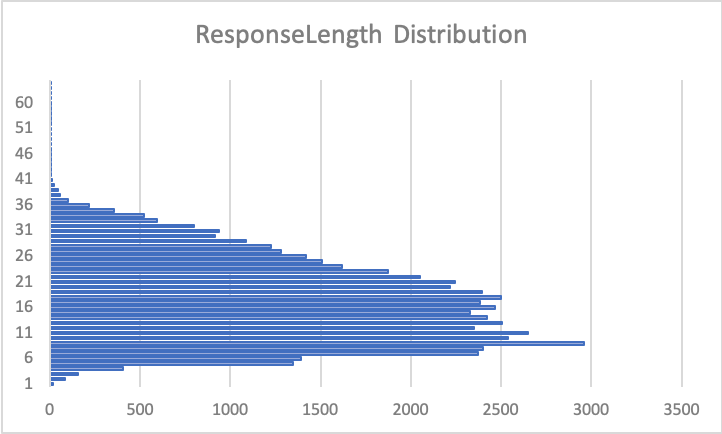}
}
\subfigure[Distribution of information load in dialogues]{
    \includegraphics[width=0.8\columnwidth]{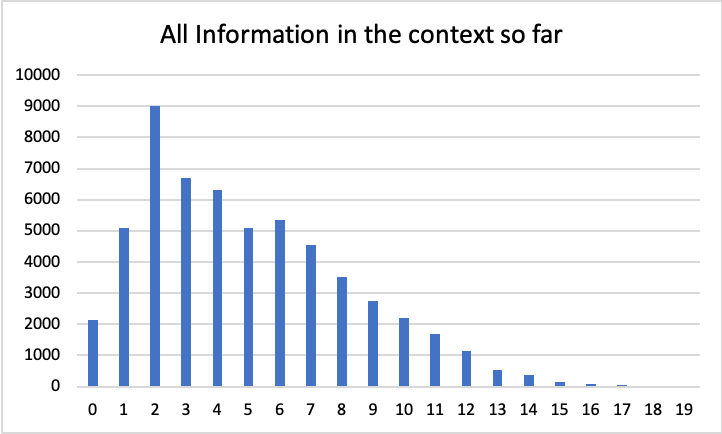}
}
\caption{Analysis of information available through probe tasks on MultiWoZ dialogue dataset.}
\label{fig:distribution-multiwoz}
\end{figure*}

Before creating the probe tasks to test the model, we wanted to ensure that the outputs of the tasks have a spread out distribution over the outputs such that the model can be evaluated on its understanding with the probe tasks. The distributional analysis on MultiWoZ data is shown in Figure \ref{fig:distribution-multiwoz} suggests that in addition to diversity in token level there is diversity in the underlying information which charcterizes the response. 

To that end, we create probe tasks with the annotated information extracted from the dataset to evaluate the understanding of the dialogue architectures.

\section{Additional Experiments}
\label{sec:addn-experiments}
Similar to the encoder representation observed on the MultiWoZ dataset, we observed the encoder representation between the first and the last epoch with training on PersonaChat dataset. 
\begin{figure*}[!h]
\centering
\subfigure[Seq2Seq Model after 0 Epoch]{
    \includegraphics[width=0.8\columnwidth]{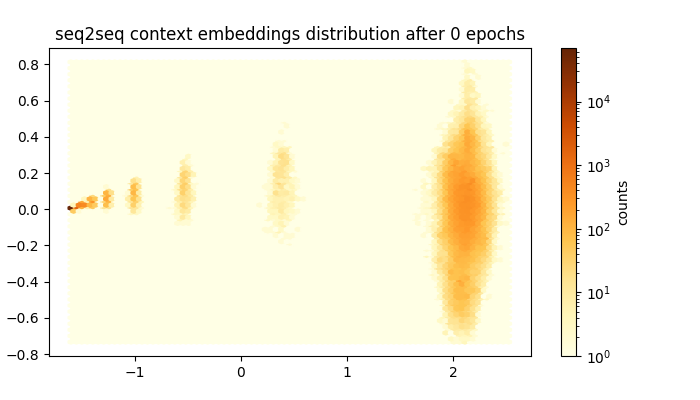}
}
\subfigure[Seq2Seq Model after Last Epoch]{
    \includegraphics[width=0.8\columnwidth]{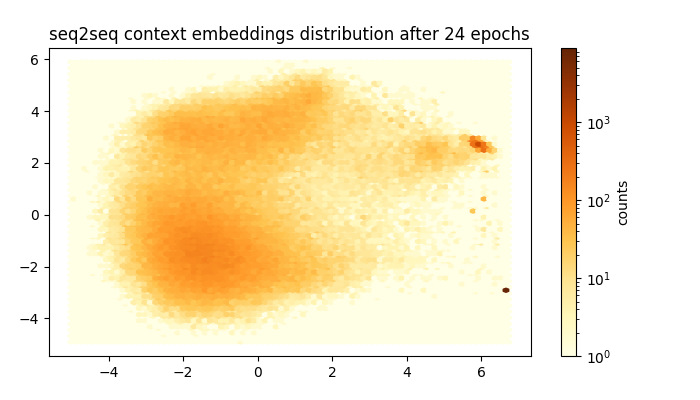}
}
\\
\subfigure[Seq2Seq Attention Model after 0 Epoch]{
    \includegraphics[width=0.8\columnwidth]{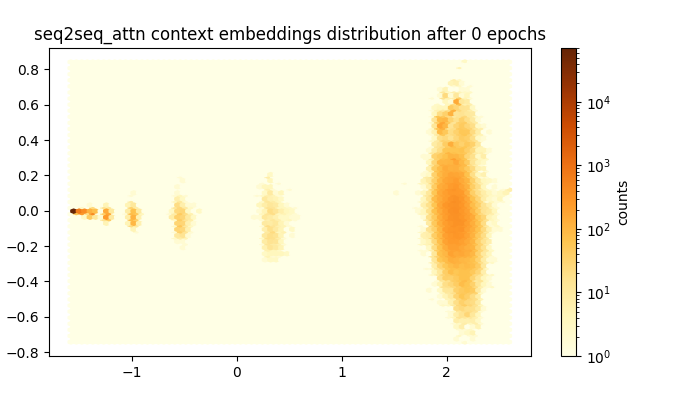}
}
\subfigure[Seq2Seq Attention Model after Last Epoch]{
    \includegraphics[width=0.8\columnwidth]{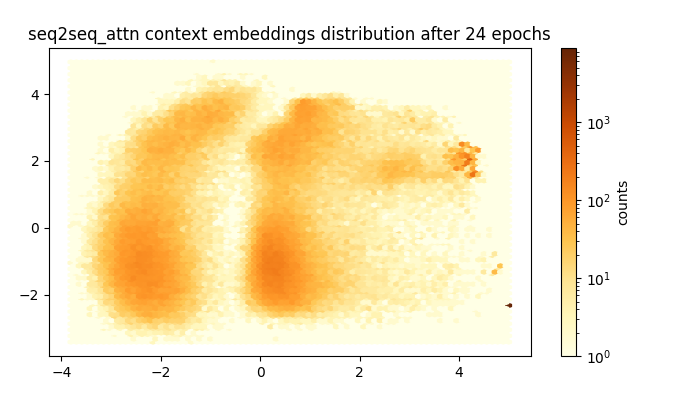}
}
\\
\subfigure[Bi-LSTM Attention Model after 0 Epoch]{
    \includegraphics[width=0.8\columnwidth]{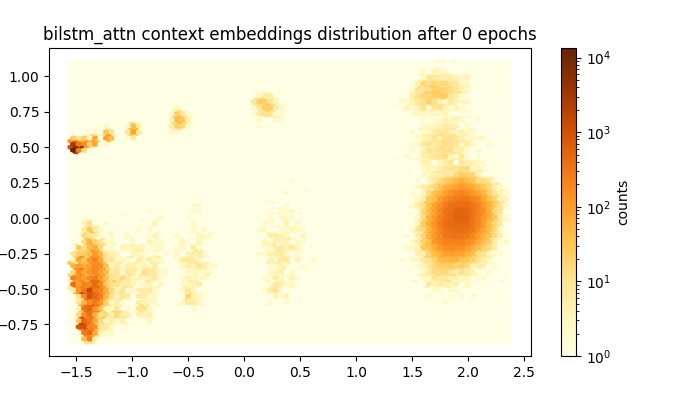}
}
\subfigure[Bi-LSTM Attention Model after Last Epoch]{
    \includegraphics[width=0.8\columnwidth]{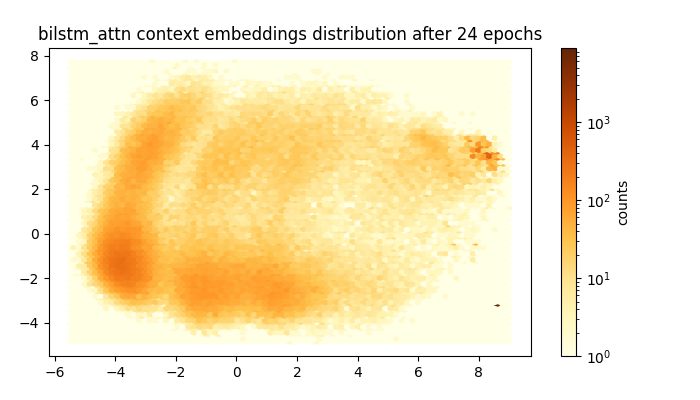}
}
\\
\subfigure[HRED Model after 0 Epoch]{
    \includegraphics[width=0.8\columnwidth]{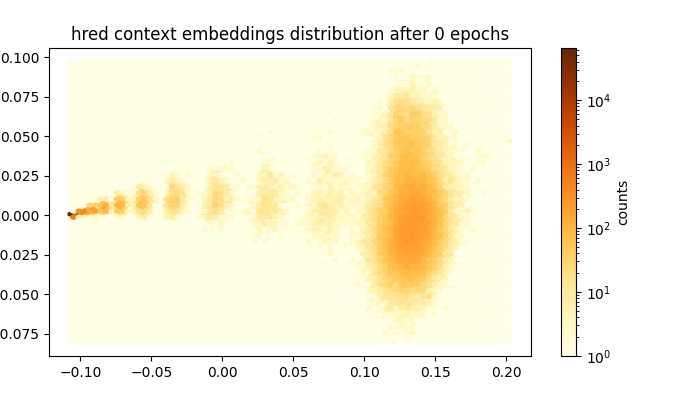}
}
\subfigure[HRED Model after Last Epoch]{
    \includegraphics[width=0.8\columnwidth]{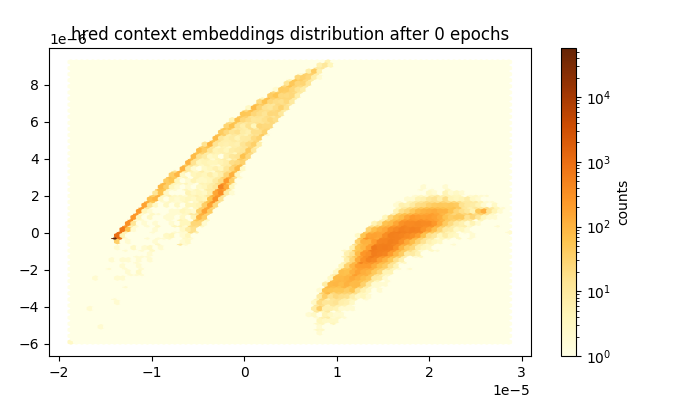}
}
\\
\subfigure[Transformer Model after 1 Epoch]{
    \includegraphics[width=0.8\columnwidth]{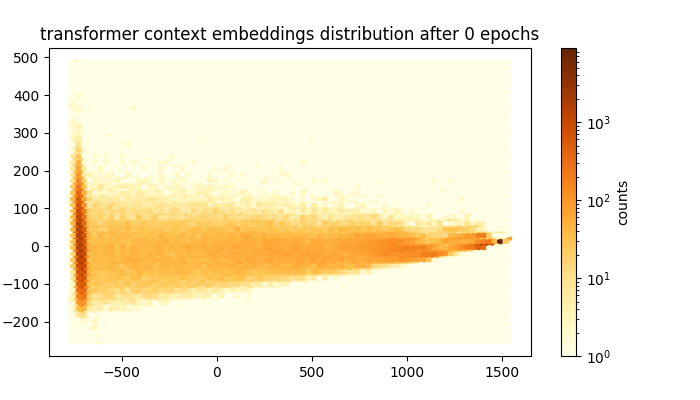}
}
\subfigure[Transformer Model after 24 Epoch]{
    \includegraphics[width=0.8\columnwidth]{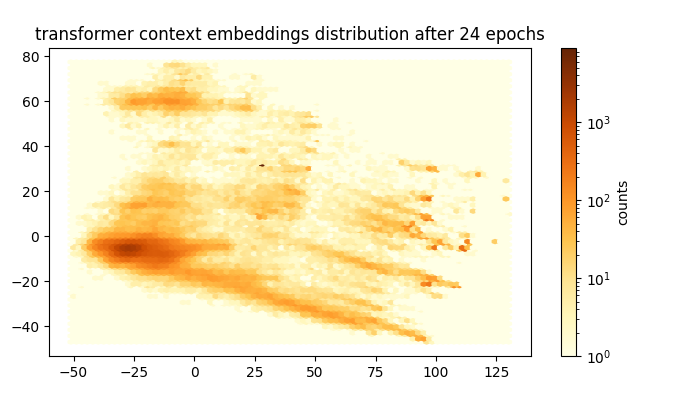}
}
\caption{Downsampled encoder hidden states on PersonaChat dataset by projecting on to two principal components. The encoder representation manifold shows that Transformer model has high capacity to encode a large dataset but not the SEQ2SEQ models.}
\label{fig:personachat-context}
\end{figure*}
PersonaChat dataset being an open-domain dialogue dataset, we did not observe strong tendency to cluster in the encoder representation. Non-existence of clusters shows that the model struggle to summarize the information available, explaining the lower performance on BLEU score and probe tasks on the dataset. 

\subsection{Comparison of Model Selection on probe tasks}

We experimented on selecting models based with METEOR, ROUGE-F1(F1), Average vector with BERT encoding as alternate selection metric to BLEU on the two datasets Table \ref{tab:multiwoz-probe-tasks-1-metric},\ref{tab:multiwoz-probe-tasks-metric},\ref{tab:persona-probe-tasks-metric}. 
\begin{table}[h!]
\tiny
\centering
      \begin{tabular}{c|a|b|a}
         \hline
         \multicolumn{4}{c}{PersonaChat Dataset}\\
         \hline
         {\bf Model} & {\bf UtteranceLoc} & {\bf WordCont} & {\bf PersonalInfo} \\
         \hline
         \rowcolor{LightBlue}\multicolumn{4}{c}{Bi-LSTM Seq2Seq + Attention} \\
         \hline
         BERT &58.89 $\pm$ 0.02 &40.43 $\pm$ 0.02 &0.00 $\pm$ 0.00    \\
         F1 & 57.27 $\pm$ 0.02 &40.28 $\pm$ 0.06 &0.02 $\pm$ 0.00   \\
         BLEU & 57.19 $\pm$ 0.05 &39.72 $\pm$ 0.08 &0.02 $\pm$ 0.00   \\
         METEOR & 57.48 $\pm$ 0.04 &39.29 $\pm$ 0.01 &0.02 $\pm$ 0.00 \\
        \hline
         \rowcolor{LightBlue}\multicolumn{4}{c}{HRED - LSTM} \\
         \hline
         BERT & 0.00 $\pm$ 0.00 &51.97 $\pm$ 0.01 &0.00 $\pm$ 0.00 \\
         F1 &0.00 $\pm$ 0.00 &51.97 $\pm$ 0.01 &0.00 $\pm$ 0.00  \\
         BLEU & 10.76 $\pm$ 3.48 &51.00 $\pm$ 0.07 &0.00 $\pm$ 0.00  \\
         METEOR & 0.00 $\pm$ 0.00 &51.97 $\pm$ 0.01 &0.00 $\pm$ 0.00 \\
         \hline
         \rowcolor{LightBlue}\multicolumn{4}{c}{LSTM Seq2Seq + Attention} \\
         \hline
         BERT & 55.12 $\pm$ 0.04 &42.04 $\pm$ 0.03 &0.00 $\pm$ 0.00  \\
         F1 & 48.11 $\pm$ 0.48 &41.84 $\pm$ 0.08 &0.01 $\pm$ 0.00  \\
         BestBLEU & 54.06 $\pm$ 0.06 &43.77 $\pm$ 0.24 &0.00 $\pm$ 0.00  \\
         METEOR & 51.91 $\pm$ 0.00 &42.36 $\pm$ 0.13 &0.00 $\pm$ 0.00 \\
         \hline
         \rowcolor{LightBlue}\multicolumn{4}{c}{LSTM Seq2Seq} \\
         \hline
         BERT & 50.32 $\pm$ 0.90 &42.99 $\pm$ 0.04 &0.00 $\pm$ 0.00   \\
         F1 & 52.23 $\pm$ 0.09 &39.98 $\pm$ 0.02 &0.01 $\pm$ 0.00    \\
         BLEU & 52.23 $\pm$ 0.08 &40.16 $\pm$ 0.04 &0.01 $\pm$ 0.00  \\
         METEOR & 52.17 $\pm$ 0.11 &40.54 $\pm$ 0.00 &0.01 $\pm$ 0.00\\
         \hline
         \rowcolor{LightBlue}\multicolumn{4}{c}{Transformer Architecture} \\
         \hline
         BERT & 42.30 $\pm$ 0.02 &45.71 $\pm$ 0.12 &0.02 $\pm$ 0.00 \\
         F1 & 40.33 $\pm$ 0.07 &47.15 $\pm$ 0.06 &0.00 $\pm$ 0.00 \\
         BLEU & 40.73 $\pm$ 0.06 &46.16 $\pm$ 0.03 &0.03 $\pm$ 0.00 \\
         METEOR & 39.73 $\pm$ 0.02 &45.76 $\pm$ 0.07 &0.01 $\pm$ 0.00 \\
         \hline
    \end{tabular}
    \caption{Comparison of models selected different selection metrics on probe tasks in PersonaChat dialogue data set. The performance is measured with \emph{F1} on the probetasks.}
    \label{tab:persona-probe-tasks-metric}
\end{table}

 \begin{table*}[h!]
        \tiny
        \centering
        \begin{tabular}{c|a|b|a|b|a|b|a|b}
         \hline
         \multicolumn{9}{c}{MultiWoZ Dataset}\\
         \hline
         {\bf Model} & {\bf UtteranceLoc} & {\bf RecentTopic} & {\bf RecentSlots} & {\bf RecentValues} & {\bf RepeatInfo} & {\bf NumRepeatInfo}& {\bf NumRecentInfo} & {\bf AllSlots} \\
         \hline
         \rowcolor{LightBlue}\multicolumn{9}{c}{LSTM Seq2Seq + Attention} \\
         \hline
         BERT &37.12 $\pm$ 2.59 &42.74 $\pm$ 16.78 &43.53 $\pm$ 4.54 &30.93 $\pm$ 0.63 &70.82 $\pm$ 0.01 &74.71 $\pm$ 0.01 &44.76 $\pm$ 1.89 &23.94 $\pm$ 6.31 \\
         F1 &58.08 $\pm$ 0.09 &89.31 $\pm$ 0.08 &66.72 $\pm$ 0.02 &39.55 $\pm$ 0.05 &71.25 $\pm$ 0.01 &75.10 $\pm$ 0.00 &62.14 $\pm$ 0.02 &52.57 $\pm$ 0.10 \\
         BLEU & 57.55 $\pm$ 0.05 &89.91 $\pm$ 0.07 &67.39 $\pm$ 0.02 &40.49 $\pm$ 0.04 &70.92 $\pm$ 0.00 &74.73 $\pm$ 0.00 &62.48 $\pm$ 0.02 &53.08 $\pm$ 0.11 \\
         METEOR &58.23 $\pm$ 0.08 &89.26 $\pm$ 0.08 &66.83 $\pm$ 0.02 &39.72 $\pm$ 0.04 &71.29 $\pm$ 0.00 &75.01 $\pm$ 0.00 &62.23 $\pm$ 0.01 &52.58 $\pm$ 0.10\\
        \hline
         \rowcolor{LightBlue}\multicolumn{9}{c}{HRED - LSTM} \\
         \hline
         BERT &18.78 $\pm$ 10.58 &23.78 $\pm$ 16.97 &16.41 $\pm$ 8.07 &10.44 $\pm$ 3.27 &71.78 $\pm$ 0.02 &75.51 $\pm$ 0.01 &19.27 $\pm$ 11.14 &13.31 $\pm$ 5.31 \\
         F1 &37.18 $\pm$ 10.38 &49.59 $\pm$ 19.55 &33.95 $\pm$ 8.98 &20.81 $\pm$ 3.26 &71.33 $\pm$ 0.01 &74.99 $\pm$ 0.01 &38.49 $\pm$ 11.14 &28.49 $\pm$ 6.63 \\
         BLEU & 37.15 $\pm$ 10.35 &50.98 $\pm$ 20.94 &34.84 $\pm$ 9.69 &20.63 $\pm$ 3.21 &71.68 $\pm$ 0.00 &75.06 $\pm$ 0.00 &38.59 $\pm$ 11.18 &30.23 $\pm$ 7.84 \\
         METEOR &41.04 $\pm$ 5.85 &50.78 $\pm$ 20.86 &44.50 $\pm$ 2.49 &28.96 $\pm$ 0.18 &71.72 $\pm$ 0.00 &75.28 $\pm$ 0.00 &50.71 $\pm$ 1.44 &30.21 $\pm$ 7.84 \\
         \hline
         \rowcolor{LightBlue}\multicolumn{9}{c}{LSTM Seq2Seq } \\
         \hline
         BERT &54.16 $\pm$ 0.94 &63.24 $\pm$ 16.20 &55.13 $\pm$ 4.78 &34.62 $\pm$ 0.76 &72.00 $\pm$ 0.00 &75.90 $\pm$ 0.00 &54.06 $\pm$ 2.46 &37.48 $\pm$ 5.21  \\
         F1 &57.56 $\pm$ 0.06 &89.44 $\pm$ 0.04 &68.00 $\pm$ 0.00 &40.98 $\pm$ 0.03 &71.22 $\pm$ 0.01 &75.32 $\pm$ 0.01 &62.78 $\pm$ 0.01 &53.07 $\pm$ 0.04\\
         BLEU& 57.37 $\pm$ 0.06 &89.45 $\pm$ 0.03 &68.08 $\pm$ 0.01 &39.78 $\pm$ 0.07 &71.28 $\pm$ 0.01 &75.36 $\pm$ 0.01 &62.33 $\pm$ 0.05 &53.40 $\pm$ 0.05 \\
         METEOR & 57.84 $\pm$ 0.04 &89.03 $\pm$ 0.01 &67.74 $\pm$ 0.01 &40.37 $\pm$ 0.10 &71.10 $\pm$ 0.00 &74.75 $\pm$ 0.00 &61.85 $\pm$ 0.00 &53.04 $\pm$ 0.02 \\
         \hline
         \rowcolor{LightBlue}\multicolumn{9}{c}{Bi-LSTM Seq2Seq + Attention} \\
         \hline
         BERT & 57.98 $\pm$ 0.03 &78.79 $\pm$ 3.19 &57.24 $\pm$ 1.71 &35.59 $\pm$ 0.34 &71.35 $\pm$ 0.00 &75.18 $\pm$ 0.01 &57.57 $\pm$ 0.18 &48.37 $\pm$ 1.11\\
         F1 &57.99 $\pm$ 0.05 &89.63 $\pm$ 0.03 &64.85 $\pm$ 0.00 &39.16 $\pm$ 0.00 &71.76 $\pm$ 0.01 &75.30 $\pm$ 0.01 &60.85 $\pm$ 0.07 &54.68 $\pm$ 0.04  \\
         BLEU & 59.04 $\pm$ 0.10 &89.85 $\pm$ 0.03 &65.03 $\pm$ 0.00 &39.06 $\pm$ 0.00 &71.98 $\pm$ 0.01 &75.63 $\pm$ 0.00 &60.36 $\pm$ 0.05 &54.96 $\pm$ 0.05  \\
         METEOR &58.45 $\pm$ 0.07 &89.28 $\pm$ 0.02 &64.21 $\pm$ 0.00 &39.19 $\pm$ 0.00 &71.54 $\pm$ 0.00 &75.35 $\pm$ 0.01 &60.49 $\pm$ 0.05 &54.65 $\pm$ 0.04  \\
         \hline
         \rowcolor{LightBlue}\multicolumn{9}{c}{Transformer Architecture} \\
         \hline
         BERT &39.11 $\pm$ 0.09 &58.38 $\pm$ 0.14 &29.97 $\pm$ 0.00 &24.50 $\pm$ 0.01 &72.39 $\pm$ 0.01 &76.02 $\pm$ 0.00 &38.80 $\pm$ 0.01 &43.37 $\pm$ 0.17 \\
         F1 & 39.89 $\pm$ 0.21 &67.44 $\pm$ 0.44 &33.37 $\pm$ 0.14 &24.96 $\pm$ 0.02 &72.75 $\pm$ 0.01 &76.26 $\pm$ 0.00 &40.43 $\pm$ 0.05 &51.19 $\pm$ 0.51\\
         BLEU &39.46 $\pm$ 0.00 &57.05 $\pm$ 1.50 &30.10 $\pm$ 0.27 &23.72 $\pm$ 0.03 &72.70 $\pm$ 0.00 &75.97 $\pm$ 0.00 &39.11 $\pm$ 0.08 &40.43 $\pm$ 1.21\\
         METEOR &38.50 $\pm$ 0.25 &56.26 $\pm$ 1.87 &30.98 $\pm$ 0.11 &24.94 $\pm$ 0.02 &72.26 $\pm$ 0.01 &75.79 $\pm$ 0.00 &39.47 $\pm$ 0.04 &38.70 $\pm$ 1.59\\
         \hline
    \end{tabular}
    \caption{Comparison of models selected different selection metrics on probe tasks in MultiWoZ dialogue data set. The performance is measured with \emph{F1} on the probetasks.}
    \label{tab:multiwoz-probe-tasks-1-metric}
\end{table*}

\begin{table*}[h!]
\tiny
\centering
    \begin{tabular}{c|a|b|a|b|a|b|a|b}
         \hline
         \multicolumn{9}{c}{MultiWoZ Dataset}\\
         \hline
         {\bf Metric} & {\bf AllValues} & {\bf NumAllInfo} & {\bf AllTopics}&{\bf NumAllTopics} & {\bf IsMultiTask} & {\bf EntitySlots} & {\bf EntityValues} & {\bf ActionSelect} \\
         \hline
         \rowcolor{LightBlue}\multicolumn{9}{c}{LSTM Seq2Seq + Attention} \\
         \hline
         BERT & 6.16 $\pm$ 0.34 &8.52 $\pm$ 2.18 &49.07 $\pm$ 5.13 &77.98 $\pm$ 0.00 &84.97 $\pm$ 0.01 &27.49 $\pm$ 1.30 &22.22 $\pm$ 0.47 &30.25 $\pm$ 6.74  \\
         F1 &12.54 $\pm$ 0.01 &26.54 $\pm$ 0.02 &75.22 $\pm$ 0.03 &79.56 $\pm$ 0.02 &84.70 $\pm$ 0.01 &41.74 $\pm$ 0.02 &31.20 $\pm$ 0.03 &60.00 $\pm$ 0.00\\
         BestBLEU  &12.81 $\pm$ 0.01 &25.73 $\pm$ 0.02 &75.33 $\pm$ 0.02 &79.39 $\pm$ 0.02 &85.30 $\pm$ 0.00 &41.29 $\pm$ 0.03 &31.57 $\pm$ 0.03 &60.14 $\pm$ 0.01\\
         METEOR & 12.53 $\pm$ 0.01 &26.62 $\pm$ 0.02 &75.21 $\pm$ 0.03 &79.52 $\pm$ 0.02 &84.67 $\pm$ 0.01 &41.70 $\pm$ 0.02 &31.48 $\pm$ 0.02 &60.06 $\pm$ 0.00 \\
        \hline
         \rowcolor{LightBlue}\multicolumn{9}{c}{HRED - LSTM} \\
         \hline
         BERT & 3.20 $\pm$ 0.31 &7.49 $\pm$ 1.68 &21.92 $\pm$ 14.41 &58.94 $\pm$ 3.05 &62.30 $\pm$ 4.46 &10.85 $\pm$ 3.53 &9.06 $\pm$ 2.46 &17.04 $\pm$ 8.72 \\
         F1 & 6.40 $\pm$ 0.32 &16.07 $\pm$ 1.97 &45.79 $\pm$ 16.07 &69.01 $\pm$ 3.62 &73.72 $\pm$ 4.79 &23.39 $\pm$ 4.22 &19.53 $\pm$ 2.87 &35.39 $\pm$ 9.73\\
         BLEU  & 6.90 $\pm$ 0.39 &14.96 $\pm$ 1.77 &46.63 $\pm$ 16.93 &68.66 $\pm$ 3.50 &72.97 $\pm$ 4.50 &24.33 $\pm$ 4.64 &19.97 $\pm$ 3.01 &35.66 $\pm$ 9.95\\
         METEOR & 6.82 $\pm$ 0.38 &15.93 $\pm$ 1.93 &54.09 $\pm$ 8.47 &79.20 $\pm$ 0.02 &85.55 $\pm$ 0.01 &30.35 $\pm$ 1.29 &25.88 $\pm$ 0.51 &36.00 $\pm$ 9.70 \\
         \hline
         \rowcolor{LightBlue}\multicolumn{9}{c}{LSTM Seq2Seq } \\
         \hline
         BERT &9.16 $\pm$ 0.27 &18.10 $\pm$ 2.47 &60.55 $\pm$ 4.58 &77.91 $\pm$ 0.03 &84.43 $\pm$ 0.02 &34.68 $\pm$ 1.90 &27.23 $\pm$ 0.79 &45.12 $\pm$ 7.11\\
         F1  & 12.92 $\pm$ 0.01 &26.47 $\pm$ 0.04 &74.63 $\pm$ 0.03 &78.44 $\pm$ 0.00 &84.05 $\pm$ 0.01 &43.66 $\pm$ 0.01 &31.83 $\pm$ 0.01 &61.11 $\pm$ 0.01 \\
         BLEU  &12.76 $\pm$ 0.01 &26.94 $\pm$ 0.04 &75.03 $\pm$ 0.03 &78.16 $\pm$ 0.00 &83.90 $\pm$ 0.00 &43.92 $\pm$ 0.01 &31.96 $\pm$ 0.01 &61.13 $\pm$ 0.00\\
         METEOR & 12.97 $\pm$ 0.00 &25.97 $\pm$ 0.03 &74.37 $\pm$ 0.01 &78.42 $\pm$ 0.00 &84.03 $\pm$ 0.01 &43.79 $\pm$ 0.04 &31.63 $\pm$ 0.02 &61.22 $\pm$ 0.02 \\
         \hline
         \rowcolor{LightBlue}\multicolumn{9}{c}{Bi-LSTM Seq2Seq + Attention} \\
         \hline
         BERT & 12.83 $\pm$ 0.10 &23.74 $\pm$ 0.13 &71.48 $\pm$ 1.07 &78.54 $\pm$ 0.07 &85.60 $\pm$ 0.00 &35.96 $\pm$ 0.72 &26.88 $\pm$ 0.07 &50.57 $\pm$ 1.36  \\
         F1 & 14.92 $\pm$ 0.00 &26.67 $\pm$ 0.07 &78.01 $\pm$ 0.01 &81.02 $\pm$ 0.06 &86.17 $\pm$ 0.00 &40.61 $\pm$ 0.00 &29.38 $\pm$ 0.01 &57.91 $\pm$ 0.01 \\
         BLEU & 15.13 $\pm$ 0.01 &25.87 $\pm$ 0.05 &78.11 $\pm$ 0.02 &80.43 $\pm$ 0.02 &86.20 $\pm$ 0.00 &40.82 $\pm$ 0.01 &29.91 $\pm$ 0.02 &57.76 $\pm$ 0.00 \\
         METEOR & 14.81 $\pm$ 0.00 &26.53 $\pm$ 0.07 &78.04 $\pm$ 0.01 &80.02 $\pm$ 0.01 &86.25 $\pm$ 0.00 &41.02 $\pm$ 0.00 &30.11 $\pm$ 0.02 &57.90 $\pm$ 0.01\\
         \hline
         \rowcolor{LightBlue}\multicolumn{9}{c}{Transformer Architecture} \\
         \hline
         BERT  & 11.81 $\pm$ 0.04 &9.01 $\pm$ 0.06 &65.01 $\pm$ 0.09 &76.23 $\pm$ 0.02 &84.38 $\pm$ 0.01 &20.60 $\pm$ 0.00 &18.87 $\pm$ 0.02 &15.48 $\pm$ 0.14 \\
         F1  & 17.97 $\pm$ 0.64 &11.26 $\pm$ 0.17 &71.08 $\pm$ 0.24 &77.82 $\pm$ 0.03 &85.27 $\pm$ 0.01 &22.47 $\pm$ 0.02 &19.06 $\pm$ 0.03 &20.24 $\pm$ 0.34 \\
         BestBLEU  & 10.43 $\pm$ 0.14 &9.71 $\pm$ 0.00 &64.42 $\pm$ 0.88 &76.10 $\pm$ 0.07 &84.20 $\pm$ 0.01 &19.83 $\pm$ 0.00 &18.34 $\pm$ 0.03 &15.35 $\pm$ 0.54\\
         METEOR & 10.77 $\pm$ 0.37 &7.92 $\pm$ 0.11 &63.64 $\pm$ 0.80 &76.58 $\pm$ 0.05 &84.50 $\pm$ 0.01 &20.17 $\pm$ 0.06 &18.38 $\pm$ 0.01 &15.03 $\pm$ 0.72\\
         \hline
    \end{tabular}
    \caption{Comparison of models selected different selection metrics on probe tasks in MultiWoZ dialogue data set. The performance is measured with \emph{F1} on the probetasks.}
    \label{tab:multiwoz-probe-tasks-metric}
\end{table*}
We compared the performance of these models on the probe tasks to understand if there is any correlation between the metric and higher performance on the probe tasks. But, we did not find any meaningful correlation between the scores and selection metric.

\subsection{Analysis of Non-linear Probing}
We experimented with multi-layer perceptron with ReLU non-linearity to analyze the trend in the model performance on the probing-tasks on the two different datasets in Tables \ref{tab:non-linear-multiwoz-probe-tasks-1},\ref{tab:non-linear-multiwoz-probe-tasks},\ref{tab:non-linear-persona-probe-tasks-metric}. Although there was some difference in the performance of individual probe tasks, the global trend of SEQ2SEQ models' performance improving with epochs while Transformer's performance decreasing with epochs remained constant. 

 \begin{table}[h!]
\tiny
\centering
      \begin{tabular}{c|a|b|a}
         \hline
         \multicolumn{4}{c}{PersonaChat Dataset}\\
         \hline
         {\bf Model} & {\bf UtteranceLoc} & {\bf WordCont} & {\bf PersonalInfo} \\
         \hline
         \rowcolor{LightBlue}\multicolumn{4}{c}{Bi-LSTM Seq2Seq + Attention} \\
         \hline
         Untrained & 38.61 $\pm$ 0.00 &39.76 $\pm$ 0.03 &0.61 $\pm$ 0.00\\
         LastEpoch & 59.05 $\pm$ 0.08 &35.82 $\pm$ 0.05 &1.29 $\pm$ 0.00  \\
         BestBLEU & 57.50 $\pm$ 0.01 &36.06 $\pm$ 0.05 &1.43 $\pm$ 0.00   \\
        \hline
         \rowcolor{LightBlue}\multicolumn{4}{c}{HRED - LSTM} \\
         \hline
         Untrained & 38.20 $\pm$ 0.03 &50.91 $\pm$ 0.08 &0.00 $\pm$ 0.00\\
         LastEpoch & 11.74 $\pm$ 4.13 &51.73 $\pm$ 0.02 &0.00 $\pm$ 0.00   \\
         BestBLEU & 12.41 $\pm$ 4.62 &51.97 $\pm$ 0.01 &0.00 $\pm$ 0.00  \\
         \hline
         \rowcolor{LightBlue}\multicolumn{4}{c}{LSTM Seq2Seq + Attention} \\
         \hline
         Untrained & 38.47 $\pm$ 1.21 &43.41 $\pm$ 0.02 &0.67 $\pm$ 0.00 \\
         LastEpoch & 52.45 $\pm$ 0.03 &34.16 $\pm$ 0.09 &1.15 $\pm$ 0.00 \\
         BestBLEU & 49.73 $\pm$ 0.23 &28.05 $\pm$ 1.78 &0.88 $\pm$ 0.01   \\
         \hline
         \rowcolor{LightBlue}\multicolumn{4}{c}{LSTM Seq2Seq} \\
         \hline
         Untrained & 44.18 $\pm$ 0.03 &43.08 $\pm$ 0.01 &0.74 $\pm$ 0.00\\
         LastEpoch & 51.23 $\pm$ 0.07 &36.12 $\pm$ 0.01 &1.07 $\pm$ 0.00  \\
         BestBLEU & 54.54 $\pm$ 0.02 &34.75 $\pm$ 0.03 &1.33 $\pm$ 0.00   \\
         \hline
         \rowcolor{LightBlue}\multicolumn{4}{c}{Transformer Architecture} \\
         \hline
         Untrained & 64.04 $\pm$ 0.01 &31.19 $\pm$ 0.77 &1.17 $\pm$ 0.00 \\
         LastEpoch & 51.74 $\pm$ 0.00 &47.99 $\pm$ 0.58 &0.00 $\pm$ 0.00 \\
         BestBLEU & 51.77 $\pm$ 0.00 &39.22 $\pm$ 4.09 &0.01 $\pm$ 0.00  \\
         \hline
    \end{tabular}
    \caption{Comparison of models on probe tasks in PersonaChat dialogue data set with a multi-layer perceptron on the different encoder representation. The performance is measured with \emph{F1} on the probetasks. The trend in the performance of the models is similar to with Logistic Regression, where the performance on WordCont task decreases with increase in training epochs, and the PersonalInfo task is still difficult for the models.}
    \label{tab:non-linear-persona-probe-tasks-metric}
\end{table}
  \begin{table*}[h!]
        \tiny
        \centering
        \begin{tabular}{c|a|b|a|b|a|b|a|b}
         \hline
         \multicolumn{9}{c}{MultiWoZ Dataset}\\
         \hline
         {\bf Model} & {\bf UtteranceLoc} & {\bf RecentTopic} & {\bf RecentSlots} & {\bf RecentValues} & {\bf RepeatInfo} & {\bf NumRepeatInfo}& {\bf NumRecentInfo} & {\bf AllSlots} \\
         \hline
         \rowcolor{LightBlue}\multicolumn{9}{c}{LSTM Seq2Seq + Attention} \\
         \hline
         Untrained &46.47 $\pm$ 0.49 &35.34 $\pm$ 0.00 &39.04 $\pm$ 0.01 &30.82 $\pm$ 0.00 &64.23 $\pm$ 0.02 &68.95 $\pm$ 0.01 &41.40 $\pm$ 0.03 &30.21 $\pm$ 0.00  \\
         LastEpoch &56.50 $\pm$ 0.06 &87.11 $\pm$ 0.01 &65.61 $\pm$ 0.00 &42.19 $\pm$ 0.00 &64.91 $\pm$ 0.00 &70.03 $\pm$ 0.01 &61.74 $\pm$ 0.06 &51.51 $\pm$ 0.01 \\
         BestBLEU &58.03 $\pm$ 0.05 &88.95 $\pm$ 0.04 &66.53 $\pm$ 0.00 &41.09 $\pm$ 0.01 &64.47 $\pm$ 0.00 &66.89 $\pm$ 0.02 &63.37 $\pm$ 0.02 &52.02 $\pm$ 0.05 \\
        \hline
         \rowcolor{LightBlue}\multicolumn{9}{c}{HRED - LSTM} \\
         \hline
         Untrained & 45.28 $\pm$ 1.56 &32.90 $\pm$ 0.02 &41.20 $\pm$ 0.03 &31.69 $\pm$ 0.02 &70.98 $\pm$ 0.01 &74.85 $\pm$ 0.01 &40.65 $\pm$ 0.03 &19.79 $\pm$ 0.03 \\
         LastEpoch & 37.96 $\pm$ 10.86 &54.21 $\pm$ 22.63 &36.27 $\pm$ 10.10 &21.27 $\pm$ 3.41 &69.42 $\pm$ 0.06 &74.02 $\pm$ 0.02 &39.45 $\pm$ 11.67 &32.77 $\pm$ 8.39 \\
         BestBLEU & 38.71 $\pm$ 11.27 &50.10 $\pm$ 20.50 &34.27 $\pm$ 9.34 &20.45 $\pm$ 3.14 &70.98 $\pm$ 0.07 &74.53 $\pm$ 0.09 &39.29 $\pm$ 11.58 &30.26 $\pm$ 7.87 \\
         \hline
         \rowcolor{LightBlue}\multicolumn{9}{c}{LSTM Seq2Seq } \\
         \hline
         Untrained &46.56 $\pm$ 0.31 &35.85 $\pm$ 0.00 &39.68 $\pm$ 0.00 &32.03 $\pm$ 0.01 &64.82 $\pm$ 0.03 &69.18 $\pm$ 0.04 &42.98 $\pm$ 0.01 &29.49 $\pm$ 0.00 \\
         LastEpoch &54.95 $\pm$ 0.06 &87.64 $\pm$ 0.01 &66.04 $\pm$ 0.01 &41.94 $\pm$ 0.01 &66.11 $\pm$ 0.00 &69.84 $\pm$ 0.01 &61.03 $\pm$ 0.04 &51.63 $\pm$ 0.01 \\
         BestBLEU&56.27 $\pm$ 0.03 &88.57 $\pm$ 0.02 &66.88 $\pm$ 0.00 &41.55 $\pm$ 0.04 &65.89 $\pm$ 0.00 &70.20 $\pm$ 0.01 &62.62 $\pm$ 0.00 &52.61 $\pm$ 0.03 \\
         \hline
         \rowcolor{LightBlue}\multicolumn{9}{c}{Bi-LSTM Seq2Seq + Attention} \\
         \hline
         Untrained &44.26 $\pm$ 0.02 &50.65 $\pm$ 0.07 &35.26 $\pm$ 0.01 &27.27 $\pm$ 0.01 &64.58 $\pm$ 0.02 &70.55 $\pm$ 0.00 &39.93 $\pm$ 0.00 &36.67 $\pm$ 0.01 \\
         LastEpoch &57.21 $\pm$ 0.04 &86.66 $\pm$ 0.01 &63.29 $\pm$ 0.01 &38.17 $\pm$ 0.02 &66.55 $\pm$ 0.00 &70.76 $\pm$ 0.00 &60.23 $\pm$ 0.07 &53.36 $\pm$ 0.00 \\
         BestBLEU &57.47 $\pm$ 0.07 &88.98 $\pm$ 0.01 &64.54 $\pm$ 0.01 &39.56 $\pm$ 0.01 &68.48 $\pm$ 0.00 &72.19 $\pm$ 0.00 &62.34 $\pm$ 0.03 &56.02 $\pm$ 0.02 \\
         \hline
         \rowcolor{LightBlue}\multicolumn{9}{c}{Transformer Architecture} \\
         \hline
         Untrained &51.24 $\pm$ 0.04 &80.25 $\pm$ 0.03 &45.58 $\pm$ 0.01 &30.61 $\pm$ 0.01 &70.36 $\pm$ 0.02 &73.25 $\pm$ 0.00 &47.47 $\pm$ 0.03 &62.94 $\pm$ 0.01 \\
         LastEpoch & 33.73 $\pm$ 0.48 &32.14 $\pm$ 1.87 &26.17 $\pm$ 1.87 &22.11 $\pm$ 1.65 &70.75 $\pm$ 0.04 &74.63 $\pm$ 0.03 &33.64 $\pm$ 3.26 &21.25 $\pm$ 0.57 \\
         BestBLEU &31.96 $\pm$ 0.47 &31.69 $\pm$ 5.29 &29.64 $\pm$ 0.34 &25.34 $\pm$ 0.21 &72.16 $\pm$ 0.00 &75.90 $\pm$ 0.00 &37.77 $\pm$ 0.37 &22.75 $\pm$ 1.44 \\
         \hline
    \end{tabular}
    \caption{The performance of multi-layer perceptron over the encoder representation of different generative dialogue models on probe tasks in MultiWoZ dialogue data set . The performance is measured with \emph{F1}. The results, similar to the probe tasks with Logistic Regression, show that SEQ2SEQ models perform significantly better than Transformer model on the probe tasks, despite the models falling behind in BLEU score.}
    \label{tab:non-linear-multiwoz-probe-tasks-1}
\end{table*}

\begin{table*}[h!]
\tiny
\centering
    \begin{tabular}{c|a|b|a|b|a|b|a|b}
         \hline
         \multicolumn{9}{c}{MultiWoZ Dataset}\\
         \hline
         {\bf Model} & {\bf AllValues} & {\bf NumAllInfo} & {\bf AllTopics}&{\bf NumAllTopics} & {\bf IsMultiTask} & {\bf EntitySlots} & {\bf EntityValues} & {\bf ActionSelect} \\
         \hline
         \rowcolor{LightBlue}\multicolumn{9}{c}{LSTM Seq2Seq + Attention} \\
         \hline
         Untrained &12.63 $\pm$ 0.01 &6.97 $\pm$ 0.01 &45.10 $\pm$ 0.00 &70.27 $\pm$ 0.02 &80.06 $\pm$ 0.01 &28.04 $\pm$ 0.00 &19.63 $\pm$ 0.00 &28.66 $\pm$ 0.00 \\
         LastEpoch & 19.33 $\pm$ 0.00 &29.28 $\pm$ 0.03 &73.35 $\pm$ 0.00 &76.25 $\pm$ 0.03 &81.54 $\pm$ 0.04 &43.48 $\pm$ 0.02 &28.42 $\pm$ 0.00 &56.21 $\pm$ 0.01 \\
         BestBLEU  &18.65 $\pm$ 0.00 &29.16 $\pm$ 0.03 &74.34 $\pm$ 0.02 &76.93 $\pm$ 0.05 &82.12 $\pm$ 0.00 &42.55 $\pm$ 0.01 &29.09 $\pm$ 0.00 &56.88 $\pm$ 0.00 \\
        \hline
         \rowcolor{LightBlue}\multicolumn{9}{c}{HRED - LSTM} \\
         \hline
         Untrained & 5.28 $\pm$ 0.00 &0.00 $\pm$ 0.00 &37.51 $\pm$ 0.04 &77.57 $\pm$ 0.00 &84.24 $\pm$ 0.00 &24.86 $\pm$ 0.07 &19.01 $\pm$ 0.04 &27.26 $\pm$ 0.01 \\
         LastEpoch &8.73 $\pm$ 0.69 &19.05 $\pm$ 2.80 &48.66 $\pm$ 18.04 &69.23 $\pm$ 3.68 &73.51 $\pm$ 4.69 &27.07 $\pm$ 5.64 &20.22 $\pm$ 3.07 &38.76 $\pm$ 11.28 \\
         BestBLEU  &8.44 $\pm$ 0.78 &18.03 $\pm$ 2.63 &46.61 $\pm$ 16.88 &68.56 $\pm$ 3.48 &73.19 $\pm$ 4.58 &24.80 $\pm$ 4.94 &20.06 $\pm$ 3.02 &34.40 $\pm$ 9.31 \\
         \hline
         \rowcolor{LightBlue}\multicolumn{9}{c}{LSTM Seq2Seq } \\
         \hline
         Untrained &13.29 $\pm$ 0.00 &6.29 $\pm$ 0.02 &43.04 $\pm$ 0.01 &73.33 $\pm$ 0.01 &80.38 $\pm$ 0.05 &27.28 $\pm$ 0.01 &20.34 $\pm$ 0.00 &29.00 $\pm$ 0.01 \\
         LastEpoch  &19.48 $\pm$ 0.00 &28.80 $\pm$ 0.01 &72.84 $\pm$ 0.01 &75.74 $\pm$ 0.01 &81.22 $\pm$ 0.00 &44.00 $\pm$ 0.01 &30.72 $\pm$ 0.01 &56.67 $\pm$ 0.01 \\
         BestBLEU  &18.78 $\pm$ 0.00 &29.67 $\pm$ 0.02 &74.28 $\pm$ 0.03 &77.08 $\pm$ 0.03 &81.89 $\pm$ 0.00 &44.21 $\pm$ 0.00 &28.88 $\pm$ 0.03 &57.16 $\pm$ 0.02 \\
         \hline
         \rowcolor{LightBlue}\multicolumn{9}{c}{Bi-LSTM Seq2Seq + Attention} \\
         \hline
         Untrained &14.94 $\pm$ 0.00 &10.88 $\pm$ 0.08 &56.79 $\pm$ 0.00 &71.36 $\pm$ 0.01 &79.49 $\pm$ 0.01 &24.21 $\pm$ 0.00 &19.04 $\pm$ 0.01 &26.07 $\pm$ 0.02 \\
         LastEpoch &20.00 $\pm$ 0.00 &28.54 $\pm$ 0.01 &74.83 $\pm$ 0.01 &78.37 $\pm$ 0.01 &83.97 $\pm$ 0.00 &42.08 $\pm$ 0.00 &29.55 $\pm$ 0.02 &55.43 $\pm$ 0.01 \\
         BestBLEU & 20.03 $\pm$ 0.01 &29.56 $\pm$ 0.04 &77.37 $\pm$ 0.00 &79.06 $\pm$ 0.01 &84.21 $\pm$ 0.01 &41.62 $\pm$ 0.03 &28.21 $\pm$ 0.00 &56.53 $\pm$ 0.00 \\
         \hline
         \rowcolor{LightBlue}\multicolumn{9}{c}{Transformer Architecture} \\
         \hline
         Untrained  & 39.62 $\pm$ 0.00 &27.26 $\pm$ 0.03 &81.16 $\pm$ 0.00 &77.63 $\pm$ 0.00 &82.79 $\pm$ 0.00 &30.29 $\pm$ 0.02 &19.72 $\pm$ 0.11 &38.48 $\pm$ 0.19 \\
         LastEpoch  &5.11 $\pm$ 0.12 &11.48 $\pm$ 0.50 &47.65 $\pm$ 1.28 &71.93 $\pm$ 0.00 &82.00 $\pm$ 0.01 &13.48 $\pm$ 0.40 &13.35 $\pm$ 0.04 &6.76 $\pm$ 0.09\\
         BestBLEU  & 5.56 $\pm$ 0.14 &7.30 $\pm$ 0.23 &50.38 $\pm$ 1.14 &73.49 $\pm$ 0.04 &81.71 $\pm$ 0.01 &23.28 $\pm$ 0.14 &12.21 $\pm$ 0.33 &7.79 $\pm$ 0.22 \\
         \hline
    \end{tabular}
    \caption{The performance of different generative dialogue models on probe tasks in MultiWoZ dialogue data set. The performance is measured with $F-1$. The results show that SEQ2SEQ models show signs of learning on the probe tasks indirectly by learning to generate next utterance. Whereas the Transformer model's performance decreased from initial to last epoch in majority of the tasks.}
    \label{tab:non-linear-multiwoz-probe-tasks}
\end{table*}
\subsection{Performance Evolution on Probe Tasks}
\begin{figure*}[h!]
\centering
\subfigure[UtteranceLoc]{
    \includegraphics[width=0.8\columnwidth]{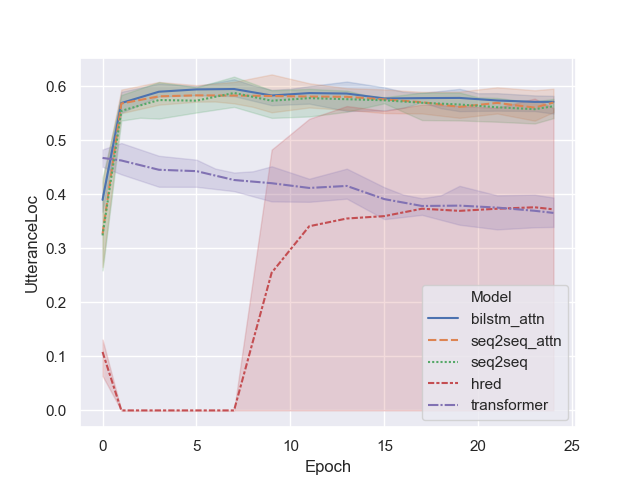}
}
\subfigure[RecentTopic]{
    \includegraphics[width=0.8\columnwidth]{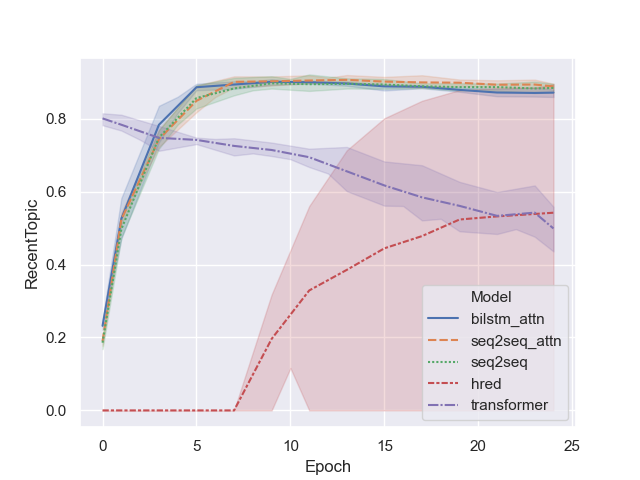}
}
\\
\subfigure[RecentSlots]{
    \includegraphics[width=0.8\columnwidth]{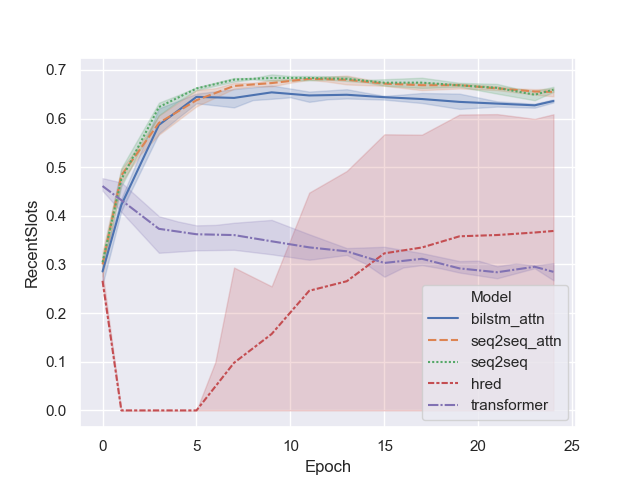}
}
\subfigure[RecentValues]{
    \includegraphics[width=0.8\columnwidth]{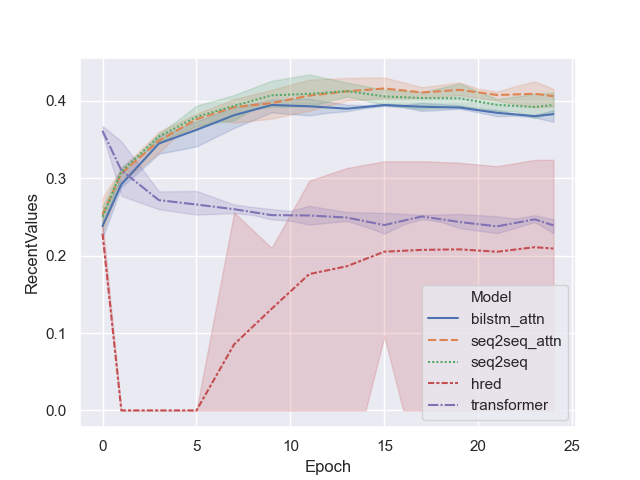}
}
\\
\subfigure[RepeatInfo]{
    \includegraphics[width=0.8\columnwidth]{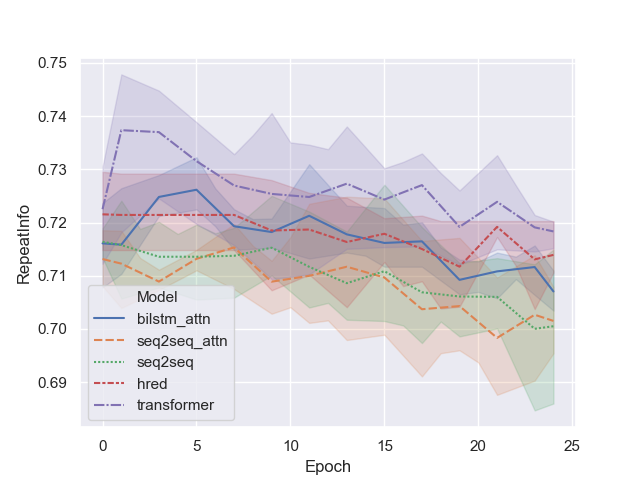}
}
\subfigure[NumRepeatInfo]{
    \includegraphics[width=0.8\columnwidth]{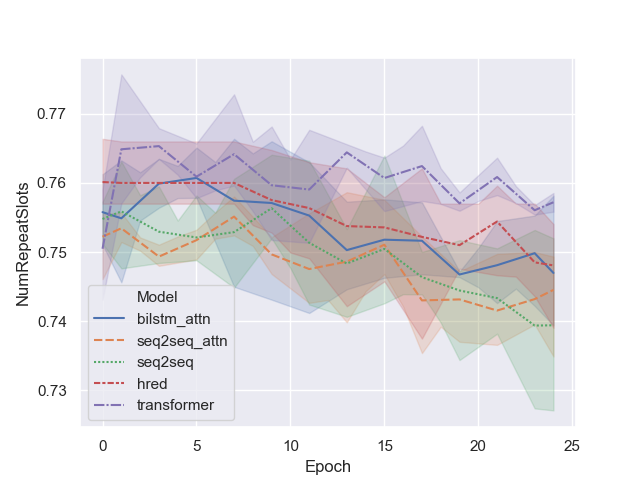}
}
\\
\subfigure[NumRecentInfo]{
    \includegraphics[width=0.8\columnwidth]{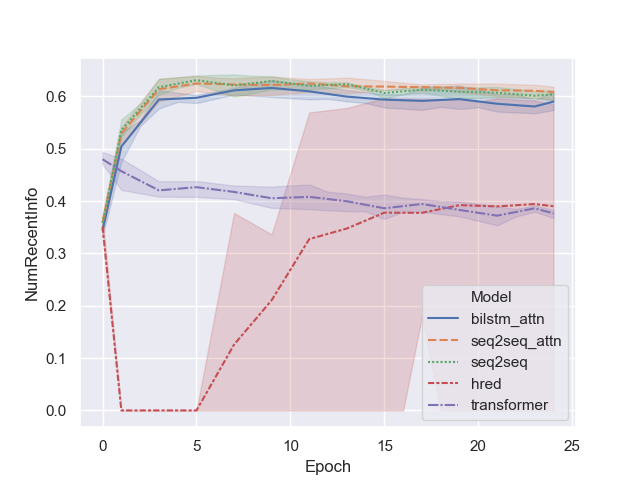}
}
\subfigure[AllSlots]{
    \includegraphics[width=0.8\columnwidth]{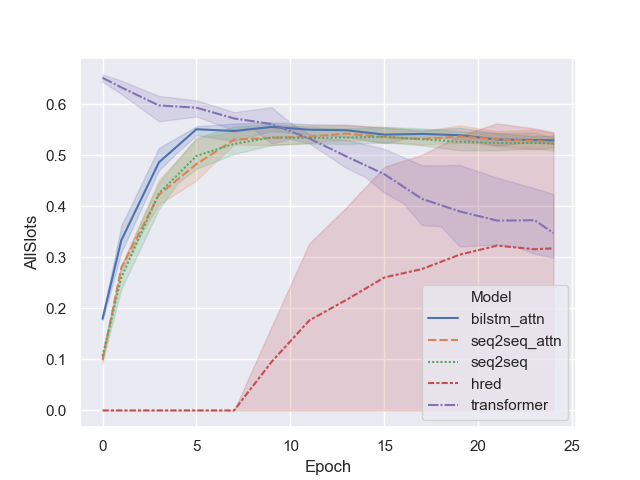}
}

\caption{Progression of performance of models on the probe tasks in MultiWoZ dataset.}
\label{fig:probetasks_table1_multiwoz}
\end{figure*}

\begin{figure*}[h!]
\centering
\subfigure[AllValues]{
    \includegraphics[width=0.8\columnwidth]{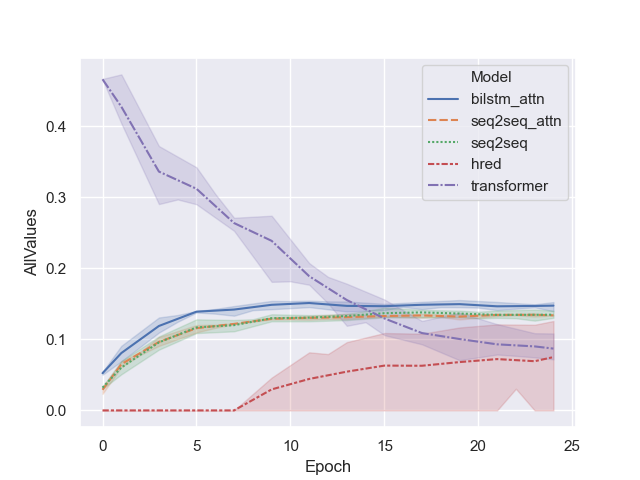}
}
\subfigure[NumAllInfo]{
    \includegraphics[width=0.8\columnwidth]{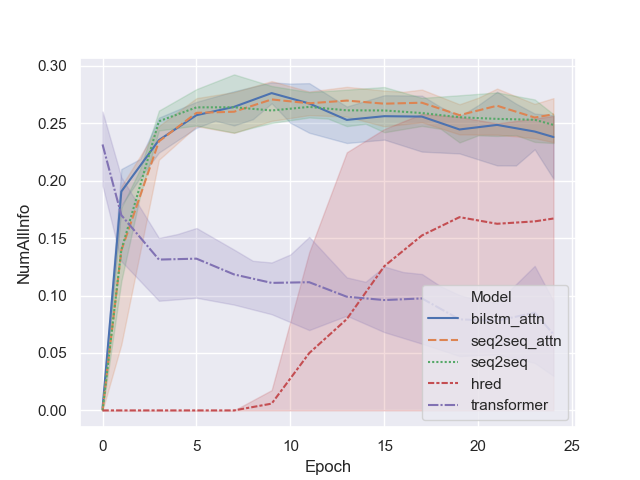}
}
\\
\subfigure[AllTopics]{
    \includegraphics[width=0.8\columnwidth]{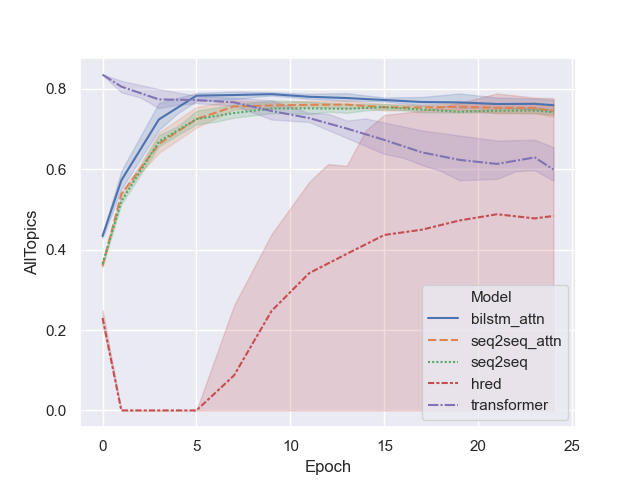}
}
\subfigure[RecentValues]{
    \includegraphics[width=0.8\columnwidth]{Images/Analysis/plot_RecentValues.png}
}
\\
\subfigure[NumAllTopics]{
    \includegraphics[width=0.8\columnwidth]{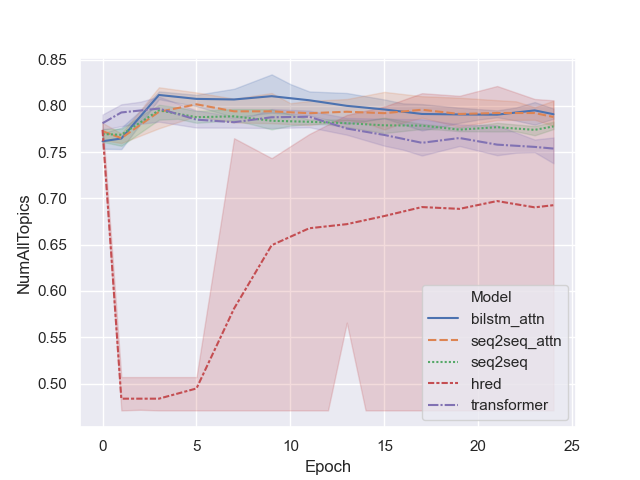}
}
\subfigure[IsMultiTask]{
    \includegraphics[width=0.8\columnwidth]{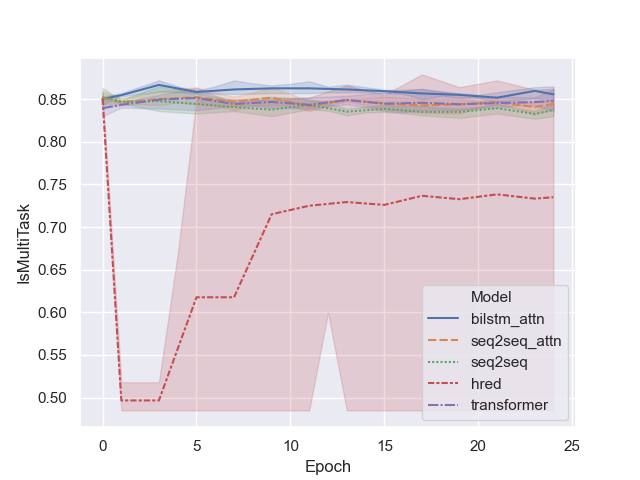}
}
\\
\subfigure[EntitySlots]{
    \includegraphics[width=0.8\columnwidth]{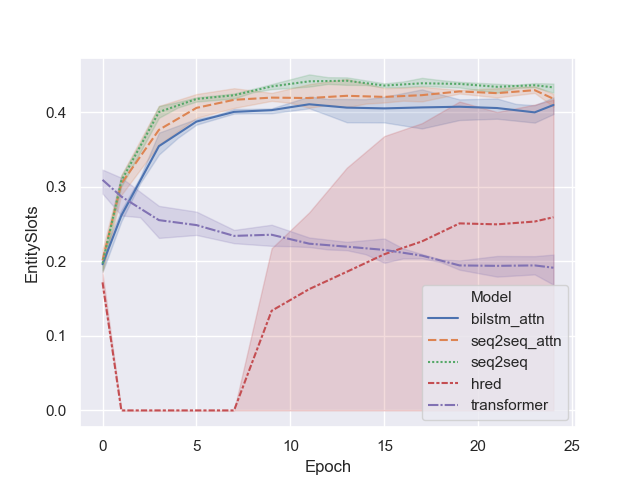}
}
\subfigure[EntityValues]{
    \includegraphics[width=0.8\columnwidth]{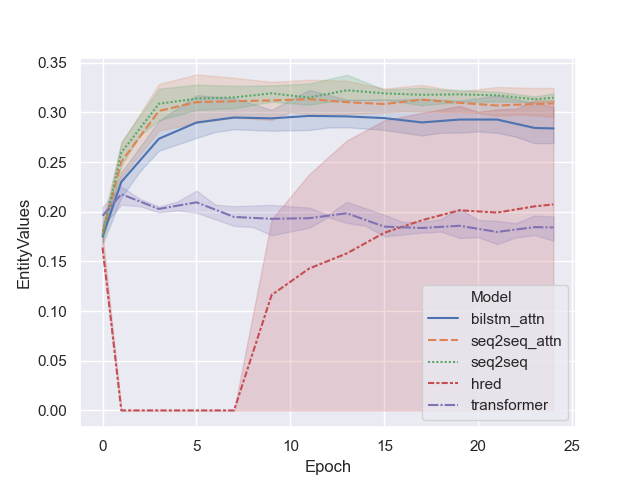}
}

\caption{Progression of performance of models on the probe tasks in MultiWoZ dataset.}
\label{fig:probetasks_table2_multiwoz}
\end{figure*}

\begin{figure*}[h!]
\centering
\subfigure[UtteranceLoc]{
    \includegraphics[width=0.8\columnwidth]{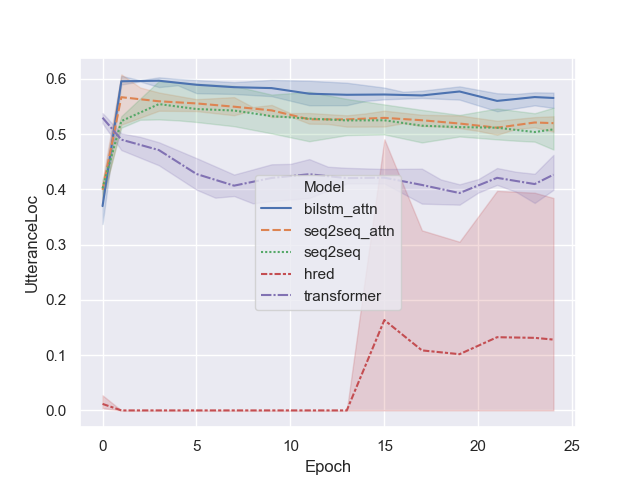}
}
\subfigure[WordCont]{
    \includegraphics[width=0.8\columnwidth]{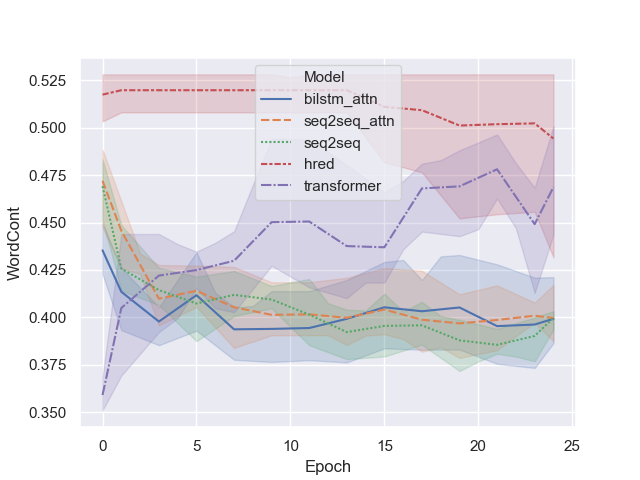}
}
\subfigure[PersonalInfo]{
    \includegraphics[width=0.8\columnwidth]{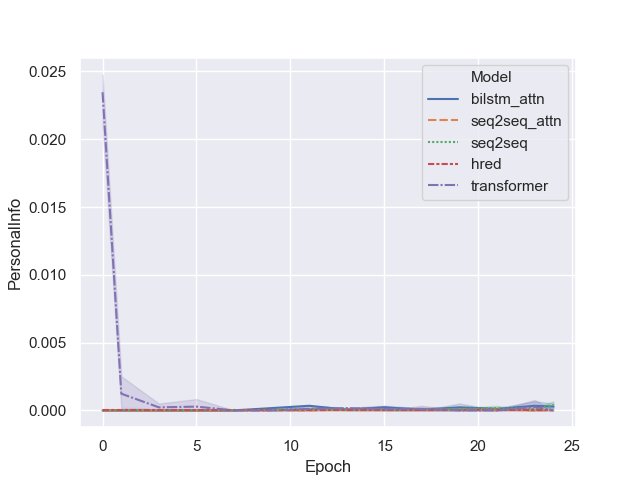}
}
\caption{Progression of performance of models on the probe tasks in PersonaChat dataset.}
\label{fig:probetasks_table1_personachat}
\end{figure*}

Although we observed in Table \ref{tab:multiwoz-probe-tasks-1}, \ref{tab:multiwoz-probe-tasks} and \ref{tab:persona-probe-tasks} that Transformer model architecture's performance decreasing while its BLEU score is higher in the task, we experimented to observe the evolution of the model performance on these probe tasks in Figures \ref{fig:probetasks_table1_multiwoz},\ref{fig:probetasks_table2_multiwoz} and \ref{fig:probetasks_table1_personachat}. We observed the trend in transformer in medium and hard probe tasks that its performance decreased and almost always stayed below the SEQ2SEQ models. This shows that the BLEU score and performance on probe tasks do not correlate, as inductive biases in a model can force the model to overfit to the patterns in text without actually understanding it.

\section{Human Evaluation Details}

\begin{figure*}[h!]
\centering
\subfigure[Example1]{
    \includegraphics[width=0.8\columnwidth]{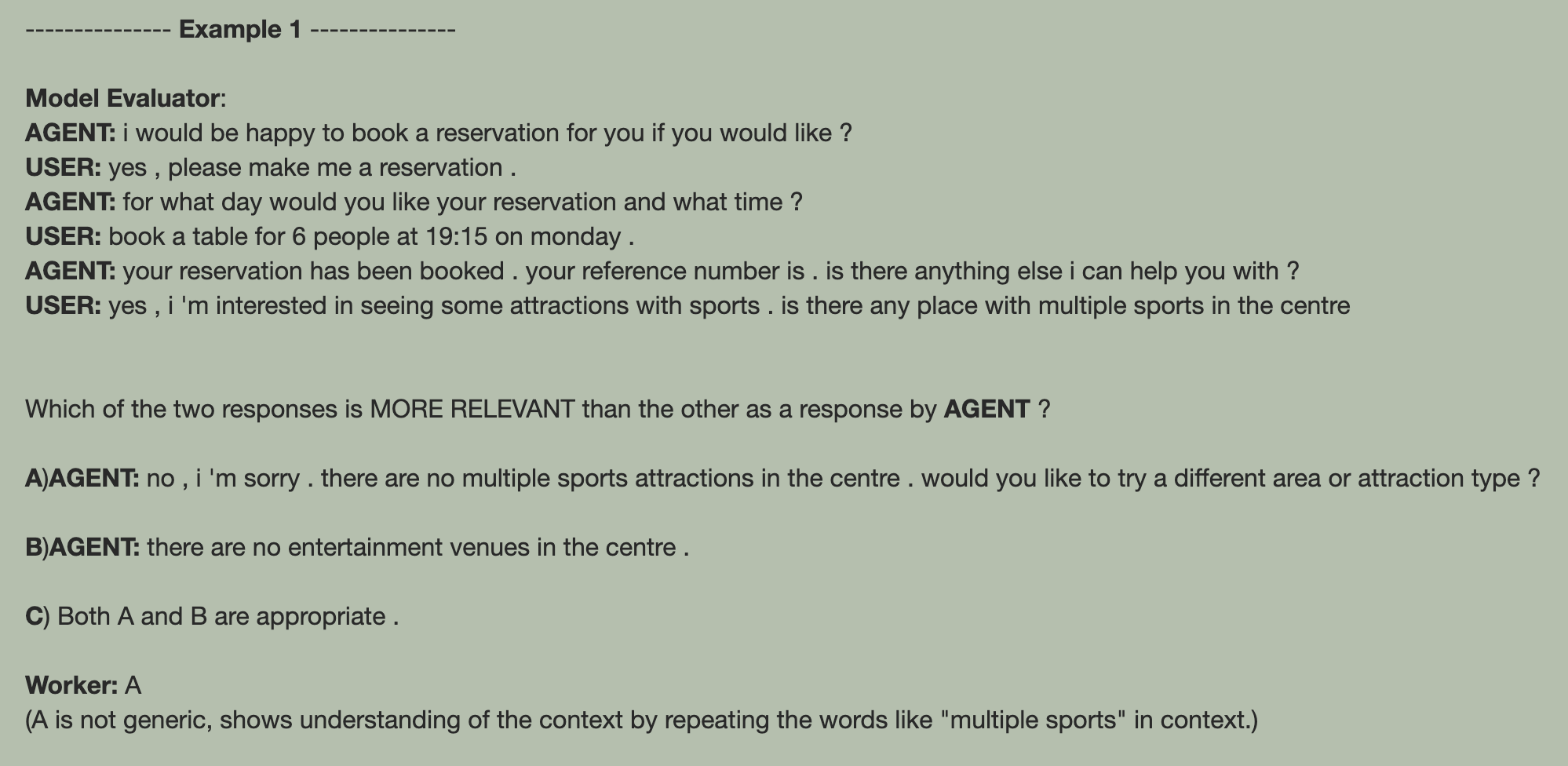}
}
\subfigure[Example2]{
    \includegraphics[width=0.8\columnwidth]{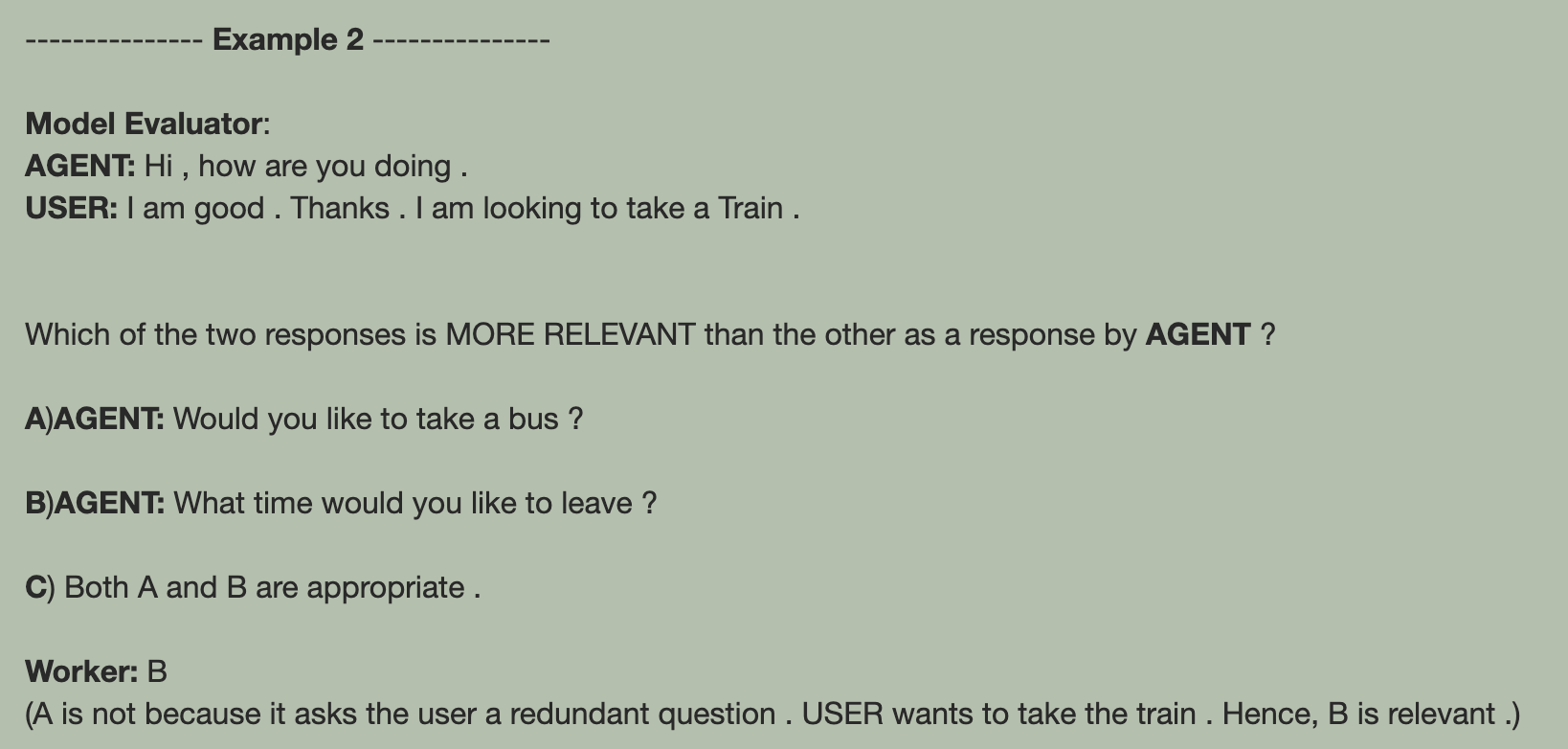}
}
\\
\subfigure[Example3]{
    \includegraphics[width=0.8\columnwidth]{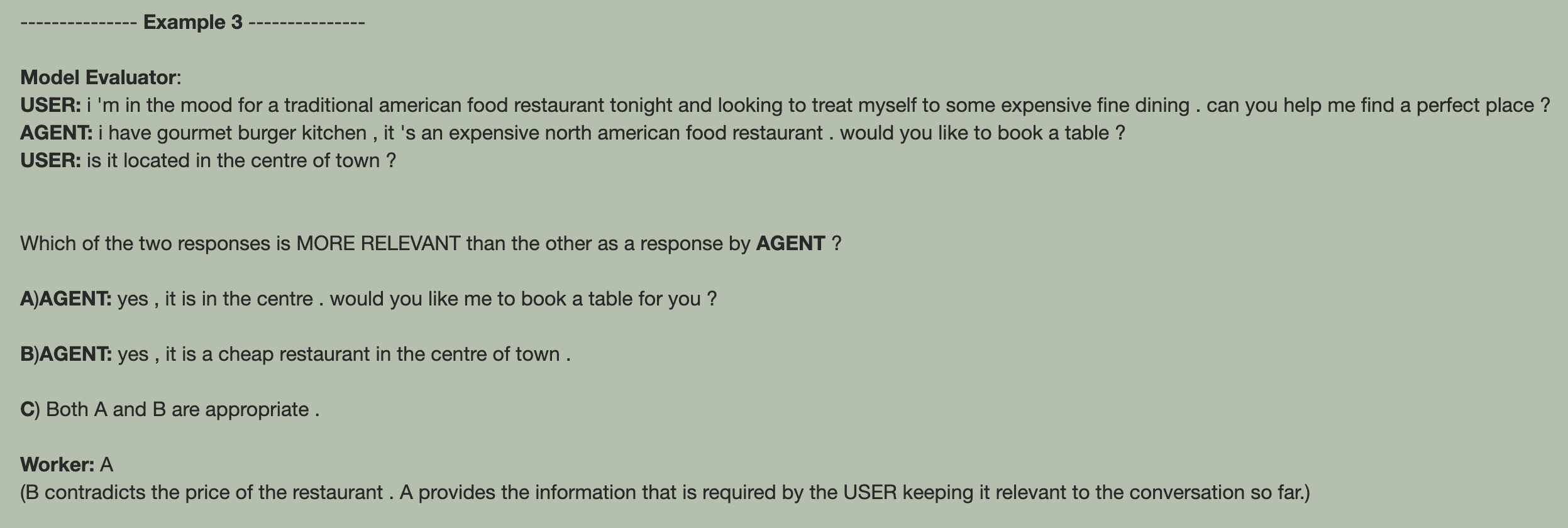}
}
\subfigure[Example4]{
    \includegraphics[width=0.8\columnwidth]{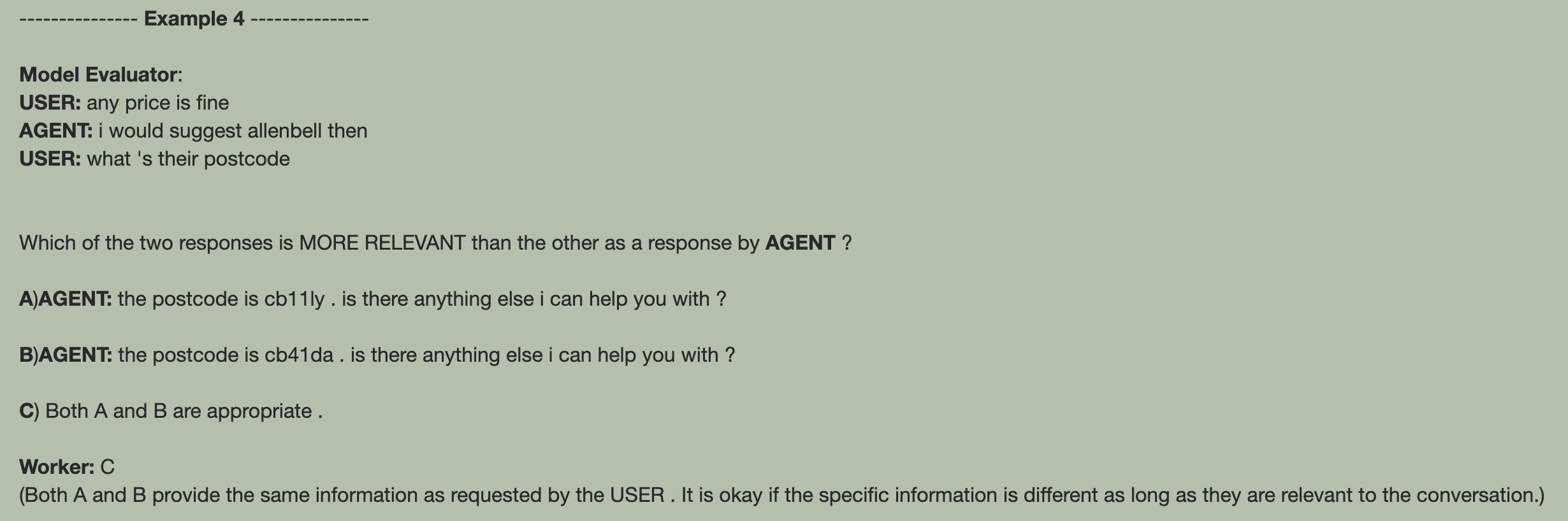}
}
\\
\subfigure[General Instruction]{
    \includegraphics[width=0.8\columnwidth]{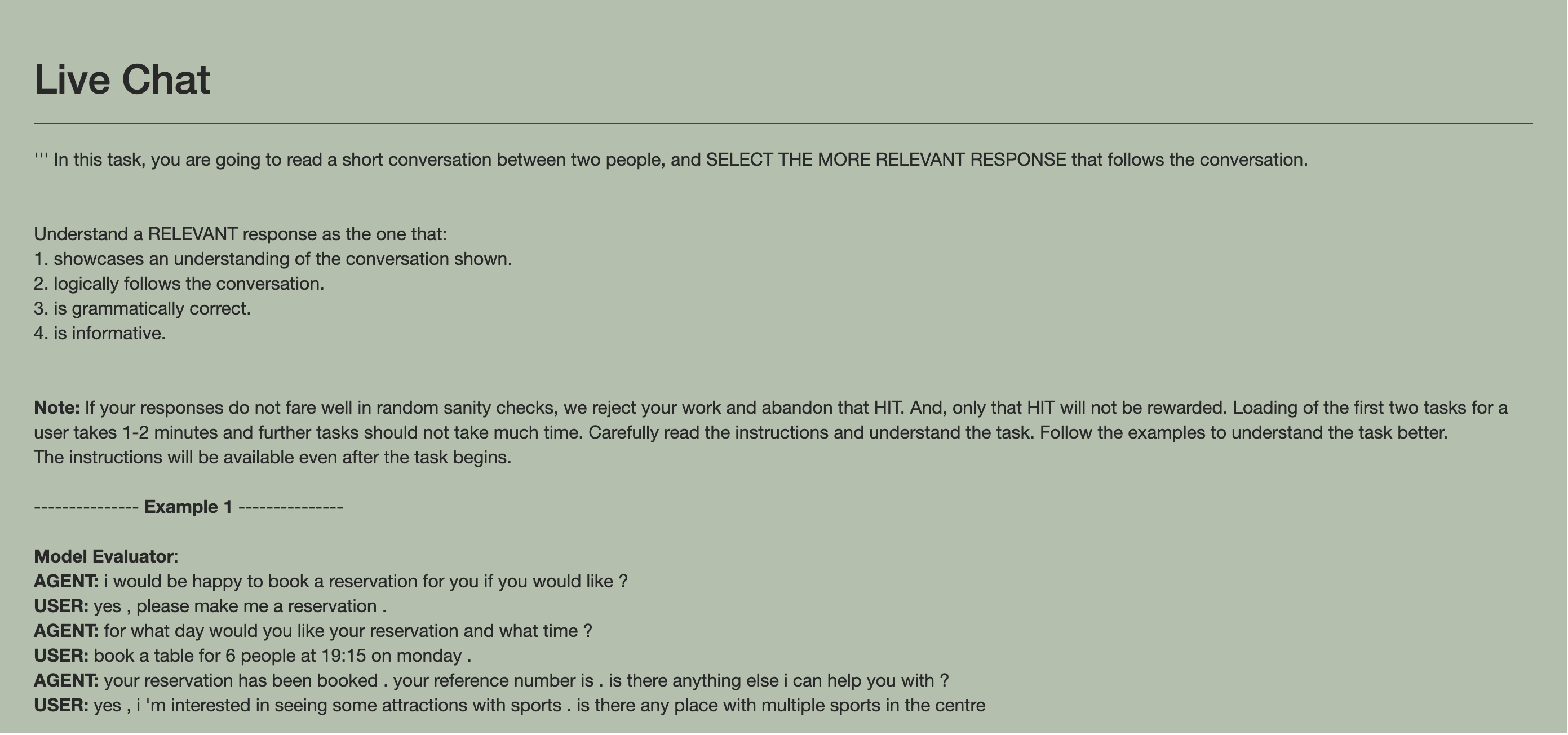}
}
\subfigure[HIT]{
    \includegraphics[width=0.8\columnwidth]{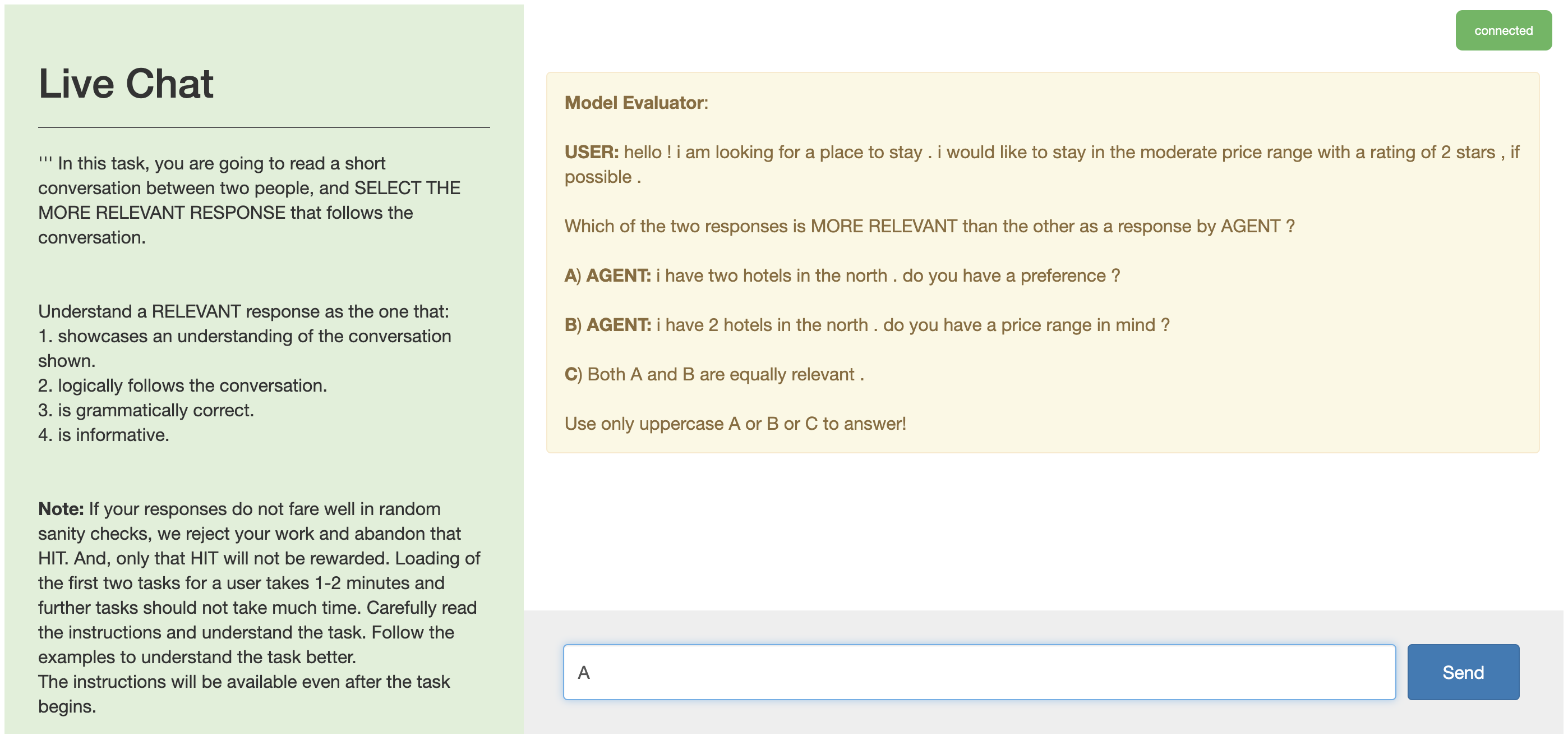}
}
\caption{Examples used to illustrate the nature of the task for data collection with ParlAI and Amazon Mechanical Turk.}
\label{fig:human_eval_examples}
\end{figure*}

We collected human annotation on comparing the relevance between two responses from the same model architecture but with different seeds. We used ParlAI's mturk framework to set up our human evaluation (Figure \ref{fig:human_eval_examples}). 

We provide the users a detailed set of instructions on relevance of the generated text with four different examples. On top of that, we start the data collection with a few sanity check questions (the participant is not told that it is a test question) where the correct answers were obvious. When the participant failed thrice, we soft block them and do not allow them to participate in our data collection. 

With this set up, we collected data only from 508 participants out of 1004 who took the sanity check task. This ensured the quality of the data collected.

\end{document}